\documentclass[lettersize,journal]{IEEEtran}
\usepackage{amsmath,amsfonts}
\usepackage{algorithmic}
\usepackage{algorithm}
\usepackage{array}
\usepackage[caption=false,font=footnotesize,labelfont=rm,textfont=rm]{subfig}
\usepackage{textcomp}
\usepackage{stfloats}
\usepackage{url}
\usepackage{verbatim}
\usepackage{graphicx}
\usepackage{cite}
\usepackage{multirow}
\usepackage{makecell}
\usepackage{setspace}
\usepackage{booktabs}
\usepackage{amssymb}
\usepackage{bbding}
\usepackage{orcidlink}
\usepackage[normalem]{ulem}
\useunder{\uline}{\ul}{}
\usepackage{hyperref}
\hypersetup{hypertex=true,
            colorlinks=true,
            linkcolor=blue,
            anchorcolor=blue,
            citecolor=blue,
            linktocpage}
\begin{document}

\title{SciceVPR: Stable Cross-Image Correlation Enhanced Model for Visual Place Recognition}

\author{Shanshan Wan$^{\orcidlink{0009-0007-8497-1582}}$, Yingmei Wei, Lai Kang$^{\ast}$, Tianrui Shen$^{\orcidlink{0000-0001-5300-4396}}$, Haixuan Wang, and Yee-Hong Yang$^{\orcidlink{0000-0002-7194-3327}}$,~\IEEEmembership{Life Senior Member,~IEEE}
\thanks{*Corresponding author:
Lai Kang.}
\thanks{Shanshan Wan, Yingmei Wei, Lai Kang, Tianrui Shen, and Haixuan Wang are with the College of Systems Engineering and the Laboratory for Big Data and Decision, National University of Defense Technology, Changsha, Hunan 410073, China (e-mail: wanshanshan16@nudt.edu.cn; weiyingmei@nudt.edu.cn; kanglai@nudt.edu.cn; shentianrui@nudt.edu.cn; wang77@nudt.edu.cn).}
\thanks{Yee-Hong Yang is with the Department of Computing Science, University of Alberta, Edmonton, AB T6G 2E9, Canada (e-mail: herberty@ualberta.ca).}}

\markboth{This work has been accepted by Neurocomputing. The final version can be accessed via \MakeLowercase{\href{https://www.sciencedirect.com/science/article/pii/S0925231225032114}{this link}.}}
{Shell \MakeLowercase{\textit{et al.}}: A Sample Article Using IEEEtran.cls for IEEE Journals}


\maketitle

\begin{abstract}
Visual Place Recognition (VPR) is a major challenge for robotics and autonomous systems, with the goal of predicting the location of an image based solely on its visual features. State-of-the-art (SOTA) models extract global descriptors using the powerful foundation model DINOv2 as backbone. These models either explore the cross-image correlation or propose a time-consuming two-stage re-ranking strategy to achieve better performance. However, existing works only utilize the final output of DINOv2, and the current cross-image correlation causes unstable retrieval results. To produce both discriminative and constant global descriptors, this paper proposes a \textbf{s}table \textbf{c}ross-\textbf{i}mage \textbf{c}orrelation \textbf{e}nhanced model for VPR called SciceVPR. This model explores the full potential of DINOv2 in providing useful feature representations that implicitly encode valuable contextual knowledge. Specifically, SciceVPR first uses a multi-layer feature fusion module to capture increasingly detailed task-relevant channel and spatial information from the multi-layer output of DINOv2. Secondly, SciceVPR considers the invariant correlation between images within a batch as valuable knowledge to be distilled into the proposed self-enhanced encoder. In this way, SciceVPR can acquire fairly robust global features regardless of domain shifts (e.g., changes in illumination, weather and viewpoint between pictures taken in the same place). Experimental results demonstrate that the base variant, SciceVPR-B, outperforms SOTA one-stage methods with single input on multiple datasets with varying domain conditions. The large variant, SciceVPR-L, performs on par with SOTA two-stage models, scoring over 3\% higher in Recall@1 compared to existing models on the challenging Tokyo24/7 dataset. Our code is available at \href{https://github.com/shuimushan/SciceVPR}{https://github.com/shuimushan/SciceVPR}.
\end{abstract}

\begin{IEEEkeywords}
Foundation model, multi-layer feature fusion, self-enhanced encoder, knowledge distillation, visual place recognition.
\end{IEEEkeywords}

\section{Introduction}
\IEEEPARstart{V}{isual} place recognition (VPR) aims at predicting the location of a query image estimated by the locations of the most visually similar images from a database \cite{netvlad}. VPR is a fundamental capability for robot state estimation \cite{anyloc} and is widely applied in mobile robot localization \cite{orb,probabilistic}, autonomous driving \cite{scalable,training}, and other areas. 

In the past decade, deep learning techniques have been successfully adapted to VPR. The database and query images are usually represented by global descriptors that describe the entire image. Then, a nearest neighbor search between query and database descriptors is performed to determine the location of the query image. Global descriptor aggregation networks are mainly composed of two parts: the backbone and the aggregation layers. Early works \cite{netvlad,openibl,loss1,loss2,structvpr,gem,pooling2,rerank1,rerank2,cosplace,eigenplaces,mixvpr,transvpr,transvlad} use convolutional neural networks (CNNs) as the backbone. Simple backbone structures like VGG16 \cite{vgg16} and ResNet50 \cite{resnet} perform well on VPR tasks. Recently, vision transformers \cite{r2former,hybrid,anyloc,dino-mix,cricavpr,selavpr} have become powerful competitors to CNNs, serving as the backbone of VPR networks, especially when using foundation models \cite{dinov2} trained on
large-scale datasets. After passing the backbone structure, images are first transformed to local features and then into compact global features through the aggregation layers.
\IEEEpubidadjcol

\begin{figure}[!t]
\centering
\includegraphics[width=3.4in]{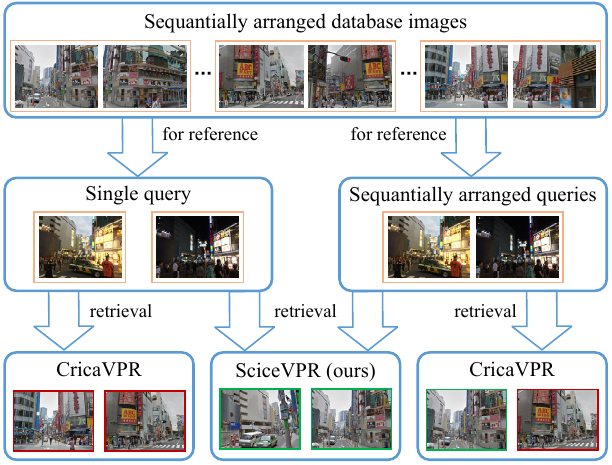}
\caption{Different retrieval results of the same query images acquired using our SciceVPR model and the state-of-the-art CricaVPR \cite{cricavpr} model to describe an image. The database images are sequentially arranged to pass through the VPR models with the batch size of 2, while we test different query situations where the number of query images is either 1 or 2. We demonstrate the most similar database images for the corresponding queries. Pictures inside an orange frame are in a batch. Red frames and green frames represent incorrect and correct retrieval results, respectively. Results show that CricaVPR produces unstable global descriptors that are affected by the number of input images, whereas our SciceVPR generates both stable and discriminative global features. }
\label{fig1}
\end{figure}

\begin{figure}[!t]
\centering
\includegraphics[width=3.4in]{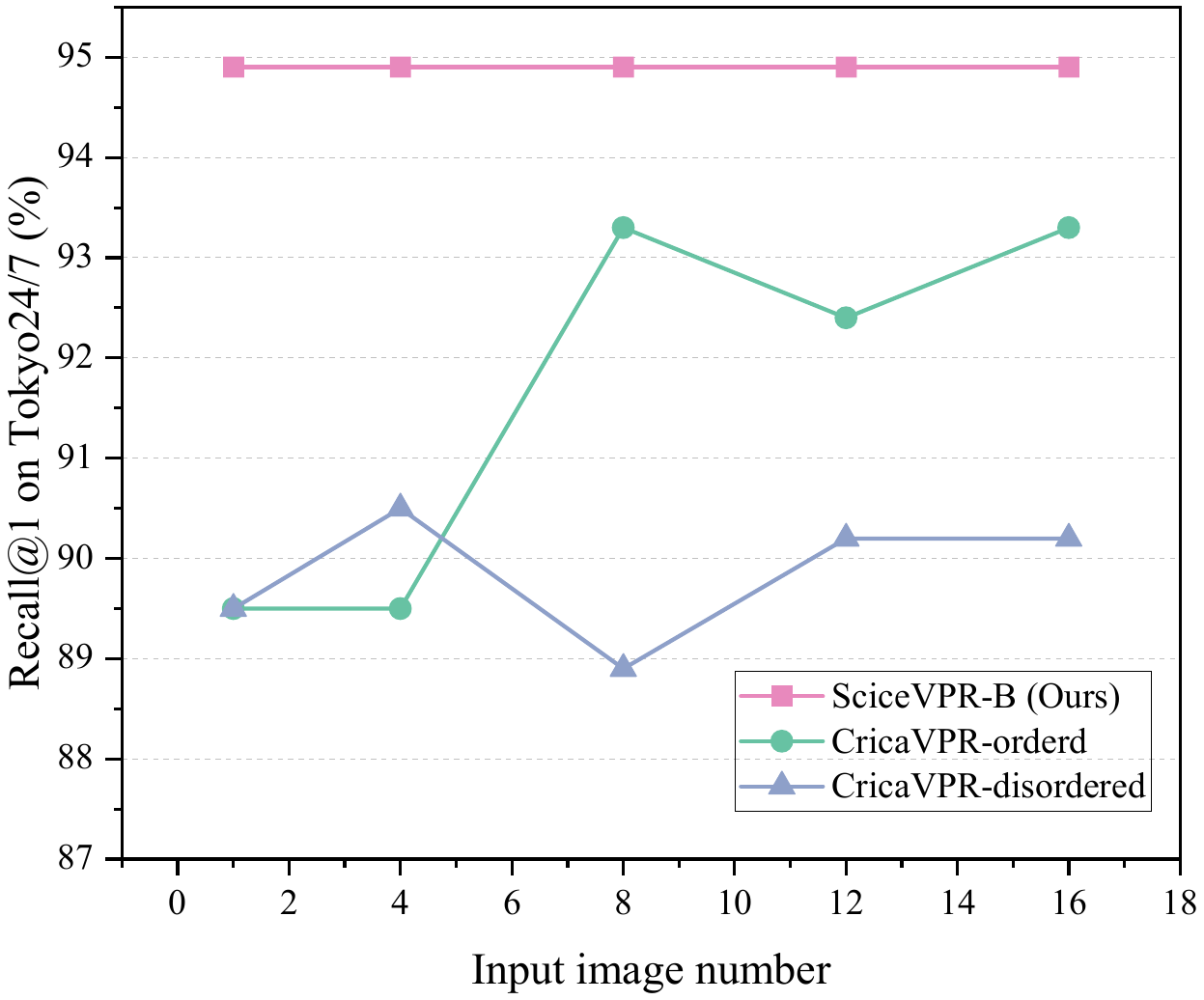}
\caption{The results of CricaVPR and SciceVPR-B models on Tokyo24/7, with descriptors' dimensionality of 10752, are compared. The database descriptors of CricaVPR are stored with a sequentially arranged input batch size 16, and its Recall@1 results vary with different query input number or orders. These results are consistently surpassed by our SciceVPR-B model.}
\label{fig2}
\end{figure}

Existing VPR networks utilize DINOv2 \cite{dinov2} as a backbone to aggregate local features from its last layer \cite{anyloc,dino-mix,cricavpr,selavpr}. However, they ignore other deep layers of DINOv2, which contain rich semantics. By concatenating patch tokens of the last 4 layers, DINOv2 achieves a significant performance boost on many dense recognition tasks compared to using features from only the last layer \cite{dinov2}. DINOv2 freezes the pre-trained backbone weights when training on downstream tasks. Similarly, we consider adopting the frozen pre-trained DINOv2 model as our backbone structure and concatenating features from its last 4 layers to get more informative local features. After obtaining the multi-layer features, we fuse them along the channel dimension and identify the spatial relationships between the local feature tokens. This approach provides task-related visual features for the successive aggregation layers.

VPR is challenging due to variations in conditions (e.g., lighting, weather, and seasonal changes), viewpoint variations, and perceptual aliasing, which can mix up similar images from different locations \cite{cricavpr}. There are three large scale datasets, which contain all the above mentioned variations: GSV-Cities \cite{gsv-cities} (0.56M images), SF-XL \cite{cosplace} (41.2M images) and MSLS \cite{msls} (1.68M images) datasets. While the MSLS dataset lacks GPS accuracy because the images are sourced from smartphone and dashcam users, GSV-Cities and SF-XL provide accurate GPS coordinates and viewing directions for each image. Many VPR networks become more robust after training on these three datasets, especially when the training datasets are GSV-Cities \cite{gsv-cities,dino-mix,mixvpr,cricavpr} or SF-XL \cite{cosplace,eigenplaces}. 

One of the SOTA models, EigenPlaces \cite{eigenplaces} explores the selection of training data from SF-XL \cite{cosplace}, which is trained on the same places from almost all possible viewpoints. Massive training samples enable EigenPlaces to perform well on VPR tasks with a simple architecture, which uses a ResNet50 backbone together with a GeM pooling layer \cite{gem}. The current best model CricaVPR \cite{cricavpr}, trained on the smaller dataset GSV-cities, uses a cross-image encoder after the multi-level GeM pooling layer to explicitly share information within a batch on a single GPU, with the goal of forcing the network to concentrate on producing invariant features. However, cross-image correlation is only effective when the input of CricaVPR is not a single image. As shown in Fig. \ref{fig1} and Fig. \ref{fig2}, the accuracy of CricaVPR depends on the number of query images and whether or not they are ordered. Hence, it is difficult, if not impossible, to apply CricaVPR in practical applications. To get both stable and discriminative global descriptors, we propose distilling cross-image invariant information into each image region within a batch using a self-enhanced encoder, which implicitly incorporates this cross-image invariant correlation. This paper presents a new \textbf{S}table \textbf{c}ross-\textbf{i}mage \textbf{c}orrelation \textbf{e}nhanced model for \textbf{VPR}, abbreviated as \textbf{SciceVPR}. Extensive experimental results show the effectiveness of our models.

The main contributions of our work include:
\begin{itemize}
\item A novel multi-layer feature fusion
module that makes use of multi-layer features from a foundation model on VPR. We adapt the features for VPR by fusing them in the channel and spatial dimensions separately.
\item A self-enhanced encoder using distilled contextual invariant knowledge, which implicitly and stably enhances the robustness of the global descriptors against challenges in VPR. To the best of our knowledge, this is the first attempt to apply knowledge distillation to handle the situation where the teacher and the student have a different number of input images.
\item Achieving state-of-the-art (SOTA) results. Extensive experiments on multiple VPR benchmark datasets show that the base variant of our model SciceVPR-B outperforms SOTA one-stage models with single input by a large margin and the large variant SciceVPR-L is on par with SOTA two-stage models.
\end{itemize}

\section{Related works}
\subsection{Visual Place Recognition}
Traditional VPR methods transform query and database images into global features using aggregation algorithms like VLAD \cite{vlad} and Bag-of-Words \cite{bow}, which aggregate hand-crafted local descriptors \cite{sift,surf}. Then a nearest neighbor search between query and database descriptors is performed to identify the location of the query image. Ever since NetVLAD \cite{netvlad} presented a trainable CNN architecture for VPR, deep learning techniques have gradually replaced traditional methods for VPR tasks. Follow-up studies continue to use CNN architectures while investigating different training strategies \cite{openibl,loss1,loss2,structvpr,cosplace,eigenplaces}, aggregation layers \cite{gem,mixvpr,pooling2}, a two-stage re-ranking method \cite{rerank1,rerank2} after global retrieval, and other approaches. Among these methods, EigenPlaces \cite{eigenplaces} ranks first by being trained on the largest VPR dataset, SF-XL \cite{cosplace}, from different points of view. EigenPlaces has a relatively simple structure, consisting of a ResNet50 backbone and a GeM pooling aggregation layer. It encodes all useful invariant information for VPR in the backbone, whose performance is greatly influenced by the training dataset. 

Gkelios et al. \cite{retrieval_vit} first propose to adopt the vision transformer (ViT) \cite{vit} for image retrieval. Subsequently, TransVPR \cite{transvpr} uses vision transformers for VPR tasks, which jointly optimizes global and patch-level features by aggregating multi-level attentions. TransVLAD \cite{transvlad} uses a CNN backbone to extract local features, which are then input to a sparse transformer encoder to efficiently encode global dependencies of these features. Wang et al. \cite{hybrid} propose a hybrid CNN-Transformer feature extraction network to get multi-level locally-global descriptors. Unlike the aforementioned VPR models, the backbone network of $R^{2}$Former \cite{r2former} is based solely on the transformer, which has been experimentally verified to outperform CNNs when used as a backbone or for providing local features for re-ranking. 

A foundation model is a model trained on a wide range of datasets and can be adapted (e.g., fine-tuned) for other downstream tasks \cite{foundation_survey,foundation_downstream1,foundation_downstream2}. Keetha et al. \cite{anyloc} investigate which of the existing foundation models \cite{dinov2,clip,dino,mae} suits VPR best. They find that DINOv2 \cite{dinov2} performs better than CLIP \cite{clip}, DINO \cite{dino} and MAE \cite{mae} on most test datasets with frozen pre-trained weights. AnyLoc \cite{anyloc} seeks to build a universal VPR solution by directly adding aggregation techniques like GeM pooling and VLAD after the frozen pre-trained backbone, without any VPR-specific training. In this way, AnyLoc can be applied to many VPR scenarios, albeit at the cost of reduced retrieval accuracy. 

Recent works \cite{dino-mix,cricavpr,selavpr} propose to fine-tune the foundation model with trainable aggregation layers. DINO-Mix \cite{dino-mix}, which is based on the original architecture of DINOv2, uses local features from patch tokens of its last layer and aggregates them through token mixer layers together with successive channel-wise and row-wise projection layers. The fine-tuning strategy of DINO-Mix is to train the last K layers (3 is reported to be the best). On the contrary, CricaVPR \cite{cricavpr} and SelaVPR \cite{selavpr} add additional adapters to the frozen DINOv2. CricaVPR directly performs multi-level GeM pooling after adapting DINOv2 and achieves the cross-image invariant correlation using a cross-image encoder. However, CricaVPR utilizes batch features as inputs for the cross-image encoder explicitly. In this way, the features passing through the encoder will only be augmented with contextual image information when the input batch size is larger than 1. Fig. \ref{fig2} shows the unstable results of CricaVPR. In order to provide both stable and discriminative global features, we propose to distill the contextual invariant correlation into a self-enhanced encoder such that our model does not depend on the number of input images. SelaVPR is the first two-stage VPR model based on a foundation model. It avoids time-consuming spatial verification in re-ranking by extracting dense local features. Nevertheless, the re-ranking stage still costs more than 4 times as long as the global matching stage and two-stage models need to store extra local features. In our proposed method, the focus is not to refine the location accuracy using an extra re-ranking strategy but to improve the network structure. Considering the effectiveness of connecting patch tokens of the last 4 layers for dense prediction tasks on DINOv2, we propose concatenating multi-layer features from DINOv2 with frozen weights and fusing task-relevant local features from both channel and spatial layers.

\begin{figure*}[!t]
\centering
\includegraphics[width=7.13in]{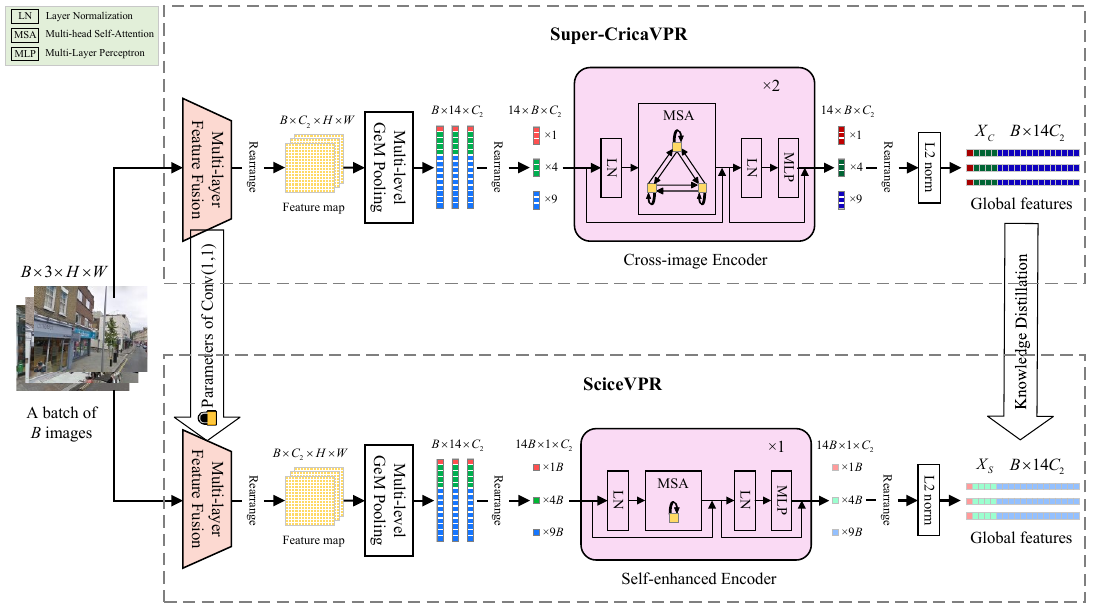}
\caption{The structure of Super-CricaVPR and SciceVPR. After training Super-CricaVPR with our proposed multi-layer feature fusion module, we use the output of Super-CricaVPR as supervision for SciceVPR. Only the parameters of conv(1,1) are passed to SciceVPR and are frozen during its training. Features are sequentially organized to pass through the cross-image encoder in Super-CricaVPR, whereas they are only augmented independently in the self-enhanced encoder of SciceVPR. We present the case where $B = 3$ and ${C_2} = 1$.}
\label{fig3}
\end{figure*}

\begin{figure}[!t]
\centering
\includegraphics[width=3.4in]{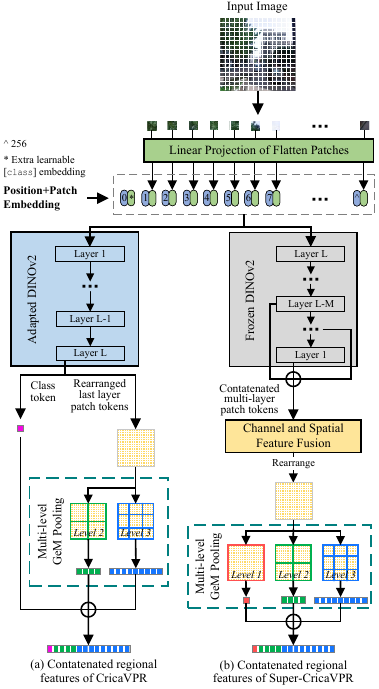}
\caption{The difference of (a) CricaVPR and (b) Super-CricaVPR in producing regional features. (a) CricaVPR only makes use of the features from the last layer of the adapted DINOv2. The output class token serves to represent the whole image and the multi-level GeM pooling is performed on the 13 regions of the rearranged patch tokens. The 14 regional features are then contatenated and passed to the cross-image encoder. (b) Super-CricaVPR makes full use of the multi-layer features from the frozen DINOv2. The contatenated multi-layer patch tokens are then fused in the channel and spatial dimensions. Similarly, multi-level GeM pooling is performed on the divided 14 regions of the rearranged patch tokens, which are then contatenated and passed to the cross-image encoder.}
\label{fig_between3_and_4}
\end{figure}

\subsection{Knowledge Distillation}
Knowledge distillation has 4 main objectives: knowledge compression, knowledge expansion, knowledge adaptation, and knowledge enhancement \cite{distillation_survey1}. The initial goal of knowledge distillation is to distill the knowledge from a larger deep neural network into a small one \cite{distill_obj11,distill_obj12,distill_obj13}, so that the compressed student network can achieve comparable performance with the teacher network but with a much lighter structure. Knowledge expansion \cite{distill_obj21,distill_obj22}, however, focuses on increasing the student's capability and generalizability beyond that of the teacher, which can be achieved by expanding the size of the student network, applying data augmentation and so on. In knowledge adaptation \cite{distill_obj31,distill_obj32}, the student network is trained on one or multiple target domains, learning from the adapted knowledge of the teacher network built on the source domains. Knowledge enhancement \cite{distill_obj41,distill_obj42} tackles the multi-task setting, where the student learns to handle different tasks under the supervision of a specialized teacher. Our distillation objective can be roughly classified as knowledge compression. However, unlike the common condition where the input for both the teacher and student networks is the same, we address the issue where the teacher accepts sequences as input while the student receives single images.

In knowledge distillation, knowledge types play an important role in student learning \cite{distillation_survey2}. There are 3 different knowledge types: response-based knowledge, feature-based knowledge, and relation-based knowledge. Response-based knowledge \cite{distill_obj11,distill_obj12,distill_obj13,distill_obj21,distill_obj22,distill_obj31,distill_obj32,distill_obj41,distill_obj42} uses the output of the teacher network to supervise the student network to make the same predictions as that of the teacher network. Feature-based knowledge uses features from the intermediate layers, i.e., feature maps, to guide the student network to produce the same features. Romero et al. \cite{distill_feature1} first propose to directly match the feature activations of the teacher and the student. Such an approach is also adopted by other researchers \cite{distill_obj13,distill_obj32}. Relation-based knowledge \cite{distill_relation1,distill_relation2} does not refer to specific layers like the previous two but instead uses the relationships between different layers or data samples. The output of VPR networks consist of global features, and we aim for our student network to produce the same contextually enhanced global features as those of the teacher network. Therefore, our knowledge type is response-based knowledge, and our distillation supervision method is more similar to the one proposed by Romero et al. \cite{distill_feature1}.  

\section{Method}
To provide stable and discriminative global features for VPR, we propose Super-CricaVPR which is based on CricaVPR \cite{cricavpr}, using our proposed multi-layer feature fusion module. Secondly, we use distillation with Super-CricaVPR as the teacher to train our student model, SciceVPR, using cross-image correlation. Fig. \ref{fig3} explains how SciceVPR is obtained.

In this section, we first briefly introduce ViT \cite{vit}. Then we present the overall structure of Super-CricaVPR, followed by details of SciceVPR, which is lighter than Super-CricaVPR but still produces stable output. Finally, we present the loss functions for training Super-CricaVPR and SciceVPR separately.

\subsection{ViT}
We adopt the DINOv2 \cite{dinov2} model as our backbone network, which is trained with ViT on large curated data without supervision. Given an input image, ViT first acquires ${x_p} \in {R^{N \times C_1}}$ patch tokens by dividing the image into $N$ flattened patches followed by linear projection of these patches to tokens. Then the tokens are concatenated with a \texttt{[class]} token to form ${x_0} = [{x_{class}};{x_p}] \in {R^{(N + 1) \times C_1}}$. Positional encoding ${E_{pos}} \in {R^{(N + 1) \times C_1}}$ is added to ${x_0}$ and the resulting embedding vector ${z_0} \in {R^{(N + 1) \times C_1}}$ is fed to the subsequent encoder. The top portion of Fig. \ref{fig_between3_and_4} shows the aforementioned procedure. The transformer encoder as depicted in Fig. \ref{fig3} contains layers mainly made of multi-head self-attention (MSA) and multi-layer perceptron (MLP) blocks. Additionally, layer norm (LN) is applied before each block and residual connections are used after each block. The above process can be summarized as:
\begin{align}
{z_0} &= {x_0} + {E_{pos}},\\
{z'_l} &= {\rm{MSA(LN}}({z_{l - 1}}){\rm{) + }}{z_{l - 1}},\\
{z_l} &= {\rm{MLP(LN(}}{{z'}_l}{\rm{)) + }}{z'_l},
\end{align}
where $l = 1, \ldots , L$.

Each layer of the transformer encoder produces $N$ patch tokens together with a class token. The patch tokens contain abundant local information in a patch and the rearranged patch tokens as a feature map (shown in Fig. \ref{fig_between3_and_4}) can be processed to describe the image. Besides, in a transformer encoder, MSA enables communication between the $N+1$ input tokens. Thus, the output class token can be directly used to represent an image after VPR-specific training. In the following section, we will discuss about the tokens we choose to use.

\begin{figure*}[!t]
\centering
\includegraphics[width=7.13in]{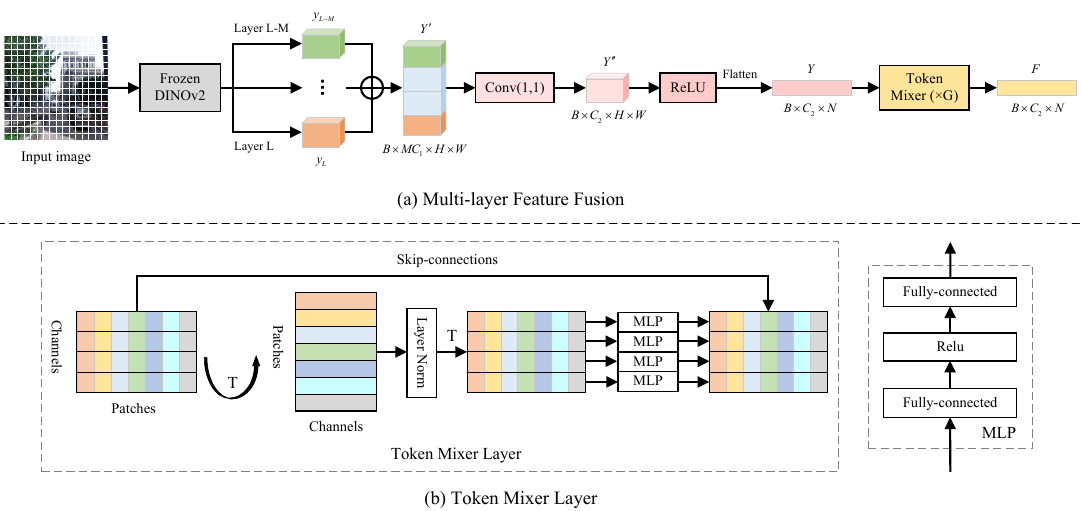}
\caption{The detailed structure of (a) the multi-layer feature fusion module. Features from the last $M$ layers of frozen DINOv2 are concatenated and fused using a channel-wise $1 \times1 $ convolution together with (b) token-wise mixer layers across all spatial token locations. }
\label{fig4}
\end{figure*}

\subsection{Super-CricaVPR}
Unlike the original CricaVPR, we directly use DINOv2 model as the backbone of Super-CricaVPR. Then we concatenate features from the last layers of DINOv2 and acquire the most task-relevant local features through a $1 \times 1$ channel-wise convolution together with token mixer layers across all spatial patch tokens. Super-CricaVPR keeps the informative representations of the foundation model and accommodates the features for VPR by incorporating exquisite architectures after the backbone instead of fine-tuning it, which is experimentally proved to have better performance than the original CricaVPR. Other part of Super-CricaVPR remains the same with CricaVPR. The overall structure of Super-CricaVPR is shown in the upper part of Fig. \ref{fig3} and the multi-layer feature fusion module is detailed in Fig. \ref{fig4}.

CricaVPR uses only the adapted DINOv2 backbone in the local feature extraction stage and directly aggregates features afterwards. Consequently, CricaVPR can utilize both the class token and the reshape of patch tokens from the final output of the adapted DINOv2 as useful extracted features. Contrary to CricaVPR, our proposed multi-layer feature fusion module adopts feature mixer components after the frozen DINOv2 backbone to make the extracted local features more relevant for VPR tasks. In our channel and spatial feature fusion module, keeping the class token will bring additional computational burden. Thus, we abandon the class token, which represents the attributes of a whole image, and focus only on the rearranged patch tokens ${y_i} \in {R^{B \times {C_1} \times H \times W}}$ from the last $M$ layers of the frozen DINOv2. Then, these channels are merged and reduced after the $1 \times 1$ convolution. ReLU is applied to introduce nonlinearity. The difference between CricaVPR and Super-CricaVPR in utilizing the output tokens from the backbone is demonstrated in Fig. \ref{fig_between3_and_4}. The corresponding formulation is as follows:
\begin{align}
{y_i} &= {\rm{Rearrange}}({\rm{LN}}({z_i})),\\
Y' &= {\rm{Concat(}}{y_{L - M}}{\rm{; }} \cdots {\rm{; }}{y_L}{\rm{)}},\\
Y &= {\rm{Flatten}}({\rm{ReLU}}({\rm{Conv}}(Y'))),
\end{align}
where $i = L-M, \ldots , L$ and $Y \in {R^{B \times {C_2} \times N}}$.

After $G$ token mixer layers \cite{mlp-mixer}, the patch tokens acquire a global reception field for VPR tasks, which allows accessing to information from different spatial locations in the image. As can be seen in Fig. \ref{fig4}, the token-mixing MLP works on each channel independently, and its parameters are shared across all channels. Each MLP consists of two fully-connected layers and a ReLU, with the output features having the same dimensions as the input features. LN is applied before each token mixer layer and residual connections are used after each layer. Each token mixer layer is given by:
\begin{align}
Y &= \{ {Y^j}\} ,\\
{F^j} &= {Y^j} + {W_2}\sigma ({W_1}{\rm{(}}LN({Y^j})),\\
F &= \{ {F^j}\},
\end{align}
where $j = 1, \ldots ,{C_2}$, $\sigma$ refers to the ReLU activation, ${W_2} \in {R^{N \times P}}$, ${W_3} \in {R^{P \times N}}$ and $F$ is the set of output feature maps of the multi-layer feature fusion module. Given the trained DINOv2 backbone, the channel diversity of $Y$ is enough for VPR tasks and we experimentally find that adding a channel mixer layer does not provide further improvement.

After gaining the instructive features $F$ from our proposed multi-layer feature fusion module, we aggregate $F$ through the multi-level GeM \cite{gem} pooling layer and get 14 ${C_2}$-dim regional features for each image. To be specific, the feature maps are separated into 3 levels ($1 \times 1$, $2 \times 2$ and $3 \times 3$) and GeM pooling is performed on these 14 local regions. Then the $x$-th regional features within a batch are sequentially arranged to pass the cross-image encoder, where the batch size of the sequences is 14 ($x = 1, \ldots , 14$) and the sequence length is the training batch size $B$. Finally, the 14 regional features of each image are concatenated and L2 normalized to form the corresponding global feature ${X_C}$.

\subsection{SciceVPR}
Super-CricaVPR utilizes the same cross-image encoder as CricaVPR. Batch features are sequentially organized to pass through the encoder where they may augment each other, which means that the output features will vary with the batch size, causing unstable global descriptors. Considering the remarkable ability of the cross-image encoder to capture the contextual invariant information within each batch of images, our SciceVPR distills the explicitly enhanced cross-image invariant knowledge into a self-enhanced encoder to produce global features that are both stable and discriminative. The architecture of SciceVPR is depicted in the lower part of Fig. \ref{fig3}.

SciceVPR consists of a multi-layer feature fusion module and a multi-level GeM pooling layer, similar to Super-CricaVPR. However, the acquired regional features are passed through the self-enhanced encoder of SciceVPR one by one. In this case, the batch size for the overall regional features is $14B$ and each regional feature is self-enhanced individually by the encoder. The self-enhanced encoder keeps the standard transformer architecture like cross-image encoder, so that it is easier to distill the contextual useful information inside the model. Besides, it needs fewer transformer layers than the cross-image encoder. Owing to the disparate data arrangement between the cross-image encoder and our self-enhanced encoder, our method reveals that knowledge distillation can also be effective when the teacher and student networks receive different amounts of input images. Finally, the self-enhanced regional features of each image are sequentially concatenated and L2 normalized to form the stable global representation ${X_S}$.

\subsection{Training Strategy}
We first train Super-CricaVPR to get a well-performing teacher model. Then, SciceVPR is trained under both the teacher and VPR supervision. The training dataset of the two models is GSV-cities \cite{gsv-cities}. The online hard mining strategy is used with multi-similarity (MS) loss \cite{ms_loss}. We apply representation learning in GSV-cities for Super-CricaVPR, similar to that of MixVPR \cite{mixvpr} and CricaVPR \cite{cricavpr}. The MS loss is calculated as follows:
\begin{align}
{L_{{\rm{MS}}}} = \frac{1}{B}\sum\limits_{q = 1}^B {\Big\{ \frac{1}{\alpha }\log \big[1 + \sum\limits_{p \in {\mathcal{P}_q}} {{e^{ - \alpha ({S_{pq}} - \lambda )}}} \big]} + \\
\frac{1}{\beta }\log \big[1 + \sum\limits_{n \in {\mathcal{N}_q}} {{e^{ - \alpha ({S_{qn}} - \lambda )}}} \big]\Big\} \nonumber,
\end{align}
where $B$ stands for the number of images in the training batch. For each query image ${I_q}$, $\mathcal{P}_q$ is the set of indices $\{ p\}$ corresponding to the positive samples for ${I_q}$, and $\mathcal{N}_q$ is the set of indices $\{ n\}$ corresponding to the negative samples for ${I_q}$. ${S_{qp}}$ and ${S_{qn}}$ are the cosine similarities of a positive pair $\{ {I_q},{I_p}\}$ and a negative pair $\{ {I_q},{I_n}\}$, respectively. $\alpha$, $\beta$ and  $\lambda$ are hyperparameters (refer to \cite{ms_loss} for more details). 

After obtaining the trained Super-CricaVPR model, we regard the contextually enhanced global features as beneficial knowledge to be passed to SciceVPR. Only the parameters of conv(1,1) in the multi-layer feature fusion module are passed and frozen from Super-CricaVPR to SciceVPR. These are closely related to the selection of powerful features from DINOv2. Since our goal is to produce self-enhanced global features 
${X_S}$ as similar to the cross-image correlation enhanced global features ${X_C}$ as possible, we use the Mean Squared Error (MSE) as the knowledge distillation loss:
\begin{equation}
{L_{{\rm{D}}}} = \frac{1}{B}\sum\limits_{i = 1}^B {{{\left\| {{X_S} - {X_C}} \right\|}^2}}.    
\end{equation}

Hence, SciceVPR is trained by minimizing the loss:
\begin{equation}
{L_{{\rm{T}}}} = \gamma {L_{{\rm{MS}}}} + \eta {L_{{\rm{D}}}},
\end{equation}
where $\gamma$ and $\eta$ are hyperparameters.

\section{Experiments}
\subsection{Implementation details}

Since the training batch size influences the results of Super-CricaVPR, we train both Super-CricaVPR and SciceVPR implemented in the Pytorch on two NVIDIA GeForce RTX 3090 GPUs with a batch size of 288 using the GSV-cities dataset, which consists of 72 places, each of which has 4 images. The resolution of the input image is 224×224 and the token dimensions of the backbone (ViT-B/14, ViT-L/14) are 768 and 1024, respectively. We connect features from the last 4 layers of the backbone networks (ViT-B/14, ViT-L/14) and reduce the feature dimension to 768. The number of token mixer layers is set to 2 with hidden dimension $16 \times 16$ and the final global features have dimension $14 \times 768$ for both Super-CricaVPR and SciceVPR. Then, PCA is performed for dimensionality reduction. We set the hyperparameters $\alpha  = 1$, $\beta  = 50$, $\lambda  = 0$ in Eq. 10, $\gamma  = 1$, $\eta  = 1$ in Eq. 12 and margin = 0.1 in online mining. We train the models using the Adam optimizer with the initial learning rate set as 0.0001 and multiplied by 0.5 after every 3 epochs, as with CricaVPR. Models with ViT-B/14 as the backbone are named Super-CricaVPR-B and SciceVPR-B, while Super-CricaVPR-L and SciceVPR-L use ViT-L/14 as the backbone. The training epoch for Super-CricaVPR is 10, while it is 2 for SciceVPR. We use the ${10^{{\rm{th}}}}$ trained model of Super-CricaVPR as the model to be distilled to SciceVPR and utilize the ${1^{{\rm{st}}}}$/${2^{{\rm{nd}}}}$ trained model of SciceVPR-B/L to be evaluated on multiple datasets. 

\begin{table}[]
\caption{Overview of the test VPR datasets\label{tab:table1}}
\centering
\footnotesize
\begin{spacing}{1.25}
\begin{tabular}{cccc}
\toprule[1pt]
Dataset        & Queries/Database & Scenery                                                   & Domain Shift  \vspace{0.25em}                                                  \\ \hline
Pitts30k-test \cite{pitts30k}  & 6816/10000       & Urban                                                     & None                                                            \\ \hline
Tokyo24/7 \cite{tokyo}      & 315/75984        & Urban                                                     & \begin{tabular}[c]{@{}c@{}}Day/Night,\vspace{-0.25em}\\ viewpoint\end{tabular}  \\ \hline
MSLS-val \cite{msls}      & 740/18871        & \begin{tabular}[c]{@{}c@{}}Urban,\vspace{-0.25em}\\ Suburban\end{tabular} & Day/Night                                                       \\ \hline
MSLS-challenge \cite{msls} & 27092/38770      & \begin{tabular}[c]{@{}c@{}}Urban,\vspace{-0.25em}\\ Suburban\end{tabular} & Long-term                                                       \\ \hline
AmsterTime \cite{amstertime}    & 1231/1231        & Urban                                                     & \begin{tabular}[c]{@{}c@{}}Long-term,\vspace{-0.25em}\\ modalities\end{tabular} \\ \hline
SVOX-Night \cite{svox}    & 823/17166        & Urban                                                     & Day/Night                                                         \\ \hline
SVOX-Overcast \cite{svox} & 872/17166        & Urban                                                     & weather                                                         \\ \hline
SVOX-Rain \cite{svox}     & 937/17166        & Urban                                                     & weather                                                         \\ \hline
SVOX-Snow \cite{svox}     & 870/17166        & Urban                                                     & weather                                                         \\ \hline
SVOX-Sun \cite{svox}      & 854/17166        & Urban                                                     & weather                                                         \\ \Xhline{1pt}
\end{tabular}
\end{spacing}
\end{table}

\subsection{Datasets and Evaluation Metric}
\noindent \textbf{Datasets.} To fully evaluate the effectiveness of SciceVPR, we conduct experiments on several VPR benchmark datasets. Their major properties are listed in Table \ref{tab:table1}.

Pitts30k \cite{pitts30k} contains 10K database images downloaded from Google Street View with GPS labels for each of the train/validation/test sets. We evaluate the models on the test set, which has 6,816 query images generated from Street View taken at different times, years from the database images. The images were captured in urban areas with diverse viewpoints.

Tokyo24/7 \cite{tokyo} is made of 75984 database images from Street View and 315 queries collected from cellphones images mainly from sidewalks, both with GPS labels. The database images are all daytime images, while the query images can be either daytime or nighttime.

MSLS \cite{msls} is a large dataset for urban and suburban VPR tasks recorded as image sequences spanning over a nine-year period. MSLS covers the following challenges: seasonal and weather changes, varying illumination throughout the day, and different viewpoints. GPS coordinates and compass angles are available for images in MSLS. We test the models on both public validation set (MSLS-val) and the withheld test set (MSLS-challenge).

AmsterTime \cite{amstertime} has 1231 pairs of images in urban areas. Each pair has one grayscale historical query image and the corresponding RGB reference image representing the same place. This dataset is challenging because of domain variations in viewpoints, modalities (RGB vs grayscale), very long-term time spans, etc.

SVOX \cite{svox} is a cross-domain dataset, which is used to evaluate VPR models on multiple weather conditions. It spans the city of Oxford, with a large database of Google Street View images; the queries, however, are from the Oxford RobotCar dataset \cite{oxford}, which provides a number of weather and illumination conditions, such as overcast, rainy, sunny, snowy and nighttime.

\vspace{0.5em}
\noindent \textbf{Evaluation metric.} We follow the evaluation metric of previous research \cite{cosplace,eigenplaces,mixvpr,cricavpr,selavpr}, where Recall@N is measured on the VPR datasets. Recall@N is defined as the percentage of queries for which at least one of the first N predictions is from the same place. For Pitts30k, Tokyo24/7, MSLS and SVOX with GPS labels, a predicted database image is considered to be from the same place as a query if their distance is within 25 meters. On the other hand, AmsterTime is a collection of images pairs, where only the counterpart of a query in the database images comes from the query’s place. In the rest of the paper, R@N refers to Recall@N.

\begin{table*}[]
\caption{Comparison to state-of-the-art methods on benchmark datasets. The best results are in \textbf{bold} and the second best {\ul underlined}. \label{tab:table2}}
\centering
\small
\begin{spacing}{1.25}
\begin{tabular}{l|c|ccc|ccc|ccc|ccc}
\Xhline{1pt}
\multirow{2}{*}{Method} & \multirow{2}{*}{Dim} & \multicolumn{3}{c|}{Tokyo24/7}                & \multicolumn{3}{c|}{Pitts30k-test}            & \multicolumn{3}{c|}{MSLS-challenge}           & \multicolumn{3}{c}{MSLS-val}                  \\ \cline{3-14} 
                        &                      & R@1           & R@5           & R@10          & R@1           & R@5           & R@10          & R@1           & R@5           & R@10          & R@1           & R@5           & R@10          \\ \hline
CosPlace \cite{cosplace}                & 512                  & 81.9          & 90.2          & 92.7          & 88.4          & 94.5          & 95.7          & 61.4          & 72.0          & 76.6          & 82.8          & 89.7          & 92.0          \\
MixVPR \cite{mixvpr}                 & 4096                 & 85.1          & 91.7          & 94.3          & 91.5          & 95.5          & 96.3          & 64.0          & 75.9          & 80.6          & 88.0          & 92.7          & 94.6          \\
EigenPlaces \cite{eigenplaces}            & 2048                 & {\ul 93.0}    & {\ul 96.2}    & {\ul 97.5}    & {\ul 92.5}    & {\ul 96.8}    & {\ul 97.6}    & 67.4          & 77.1          & 81.7          & 89.1          & 93.8          & 95.0          \\
CricaVPR-single \cite{cricavpr}        & 4096                 & 89.8          & 93.7          & 96.2          & 91.7          & 95.8          & 96.9          & {\ul 67.5}    & {\ul 79.5}    & {\ul 82.6}    & {\ul 89.2}    & \textbf{95.3} & {\ul 95.7}    \\ \hline
SciceVPR-B(ours)        & 4096                 & \textbf{94.9} & \textbf{97.8} & \textbf{98.4} & \textbf{92.9} & \textbf{96.9} & \textbf{98.0} & \textbf{69.2} & \textbf{84.3} & \textbf{87.9} & \textbf{89.3} & {\ul 95.0}          & \textbf{96.5} \\ \hline
TransVPR \cite{transvpr}               & /                    & 79.0          & 82.2          & 85.1          & 89.0          & 94.9          & 96.2          & 63.9          & 74.0          & 77.5          & 86.8          & 91.2          & 92.4          \\
StructVPR \cite{structvpr}              & /                    & -             & -             & -             & 90.3          & 96.0          & 97.3    & 69.4          & 81.5          & 85.6          & 88.4          & 94.3          & 95.0          \\
$R^{2}$Former \cite{r2former}               & /                    & 88.6          & 91.4          & 91.7          & 91.1          & 95.2          & 96.3          & 73.0          & 85.9          & 88.8          & 89.7          & 95.0          & 96.2          \\
SelaVPR \cite{selavpr}                & /                    & {\ul 94.0}    & {\ul 96.8}    & {\ul 97.5}    & {\ul 92.8}    & {\ul 96.8}    & {\ul 97.7}    & {\ul 73.5}    & \textbf{87.5} & \textbf{90.6} & \textbf{90.8} & \textbf{96.4} & \textbf{97.2} \\ \hline
SciceVPR-L(ours)        & 4096                 & \textbf{97.1} & \textbf{98.1} & \textbf{98.1} & \textbf{93.4} & \textbf{96.9} & \textbf{98.0} & \textbf{74.3} & {\ul 86.6}    & \textbf{90.6} & {\ul 90.7}    & {\ul 95.9}    & {\ul 96.8}    \\ \Xhline{1pt}
\end{tabular}
\end{spacing}
\end{table*}

\begin{table*}[]
\caption{Comparison to state-of-the-art methods on more challenging datasets. The best results are in \textbf{bold} and the second best {\ul underlined}. \label{tab:table3}}
\centering
\footnotesize
\setlength{\tabcolsep}{3pt}
\begin{spacing}{1.25}
\begin{tabular}{c|ccc|ccc|ccc|ccc|ccc|ccc}
\Xhline{1pt}
\multirow{2}{*}{Method} & \multicolumn{3}{c|}{AmsterTime}               & \multicolumn{3}{c|}{SVOX-Night}               & \multicolumn{3}{c|}{SVOX-Overcast}            & \multicolumn{3}{c|}{SVOX-Rain}                & \multicolumn{3}{c|}{SVOX-Snow}                & \multicolumn{3}{c}{SVOX-Sun}                  \\ \cline{2-19} 
                        & R@1           & R@5           & R@10          & R@1           & R@5           & R@10          & R@1           & R@5           & R@10          & R@1           & R@5           & R@10          & R@1           & R@5           & R@10          & R@1           & R@5           & R@10          \\ \hline
CosPlace \cite{cosplace}               & 38.7          & 61.3          & 67.3          & 44.8          & 63.5          & 70.0          & 88.5          & 93.9          & 95.2          & 85.2          & 91.7          & 93.8          & 89.0          & 94.0          & 94.6          & 67.3          & 79.2          & 83.8          \\
MixVPR \cite{mixvpr}                 & 40.2          & 59.1          & 64.6          & 64.4          & 79.2          & 83.1          & 96.2          & 98.3          & 98.9          & 91.5          & 97.2          & 98.1          & {\ul 96.8}    & 98.4          & 98.9          & 84.8          & 93.2          & 94.7          \\
EigenPlaces \cite{eigenplaces}            & 48.9          & 69.5          & 76.0          & 58.9          & 76.9          & 82.6          & 93.1          & 97.8          & 98.3          & 90.0          & 96.4          & 98.0          & 93.1          & 97.6          & 98.2          & 86.4          & 95.0          & 96.4          \\
CricaVPR-single \cite{cricavpr}        & 49.4          & 70.3          & 76.7          & 76.8          & 88.0          & 92.3          & 96.3          & 98.3          & 98.5          & 93.5          & 98.5          & 99.0          & 95.4          & 98.9          & 99.3          & 87.8          & 97.2          & 97.9          \\ \hline
SciceVPR-B(ours)        & {\ul 58.8}    & {\ul 82.0}    & {\ul 85.2}    & {\ul 88.3}    & {\ul 97.0}    & {\ul 98.4}    & {\ul 97.0}    & {\ul 99.0}    & {\ul 99.2}    & {\ul 96.4}    & {\ul 98.8}    & {\ul 99.1}    & 96.6          & {\ul 99.3}    & {\ul 99.5}    & {\ul 94.4}    & {\ul 98.5}    & {\ul 98.9}    \\
SciceVPR-L(ours)        & \textbf{63.0} & \textbf{83.4} & \textbf{88.2} & \textbf{94.7} & \textbf{98.5} & \textbf{99.1} & \textbf{97.9} & \textbf{99.1} & \textbf{99.7} & \textbf{97.7} & \textbf{99.4} & \textbf{99.7} & \textbf{98.7} & \textbf{99.4} & \textbf{99.7} & \textbf{95.6} & \textbf{99.1} & \textbf{99.5} \\ \Xhline{1pt}
\end{tabular}
\end{spacing}
\end{table*}

\subsection{Comparison with Previous Work}
\noindent \textbf{Baselines.} We compare SciceVPR with several SOTA VPR models. For the basic global-retrieval-based models, we choose the latest proposed CosPlace \cite{cosplace}, MixVPR \cite{mixvpr}, EigenPlaces \cite{eigenplaces} and CricaVPR \cite{cricavpr} models. MixVPR and CricaVPR are also trained on GSV-cities \cite{gsv-cities}. In the interest of fairness, we set the inference batch size to 1 for CricaVPR, meaning that the number of input images is 1 for generating both database and query descriptors. We refer to the setup as CricaVPR-single in the comparisons. Conversely, CosPlace and EigenPlaces are both trained on the largest VPR dataset SF-XL. Additionally, we also compare our model with the latest proposed two-stage models TransVPR \cite{transvpr}, StructVPR \cite{structvpr}, $R^{2}$Former \cite{r2former} and SelaVPR \cite{selavpr}. TransVPR and StructVPR are trained on two datasets Pitts30k-train \cite{pitts30k} and MSLS-train \cite{msls}, and evaluated on Pitts30k-test (or Tokyo24/7 \cite{tokyo}) and MSLS-val/challenge, respectively. Besides, $R^{2}$Former and SelaVPR are first trained on MSLS-train and tested on MSLS-val/challenge, and further fine-tuned on Pitts30k-train and tested on Pitts30k-test and Tokyo24/7.
\vspace{0.5em}

\noindent \textbf{Discussion of results.} We compare SciceVPR-B, having a backbone similar to that of CricaVPR-single, with the SOTA one-stage models. As can be seen in Table \ref{tab:table2}, our SciceVPR-B ranks first in Recall@1 and Recall@10 on all the datasets. EigenPlaces and CricaVPR-single perform quite well on the listed datasets. The former utilizes the largest training set, while the latter has a similar architecture to that of SciceVPR-B. However, SciceVPR-B has a better performance, in particular on the Tokyo24/7 dataset with a Recall@1 of 94.9\%, showing the model’s robustness to lighting (day/night) and viewpoint changes. Moreover, SciceVPR-B outperforms the compared models on the MSLS-val and the MSLS-challenge test sets, which include more long-term VPR challenges and some suburban areas lacking landmarks, demonstrating its effectiveness.

\begin{figure*}[!t]
\centering
\includegraphics[width=6.5in]{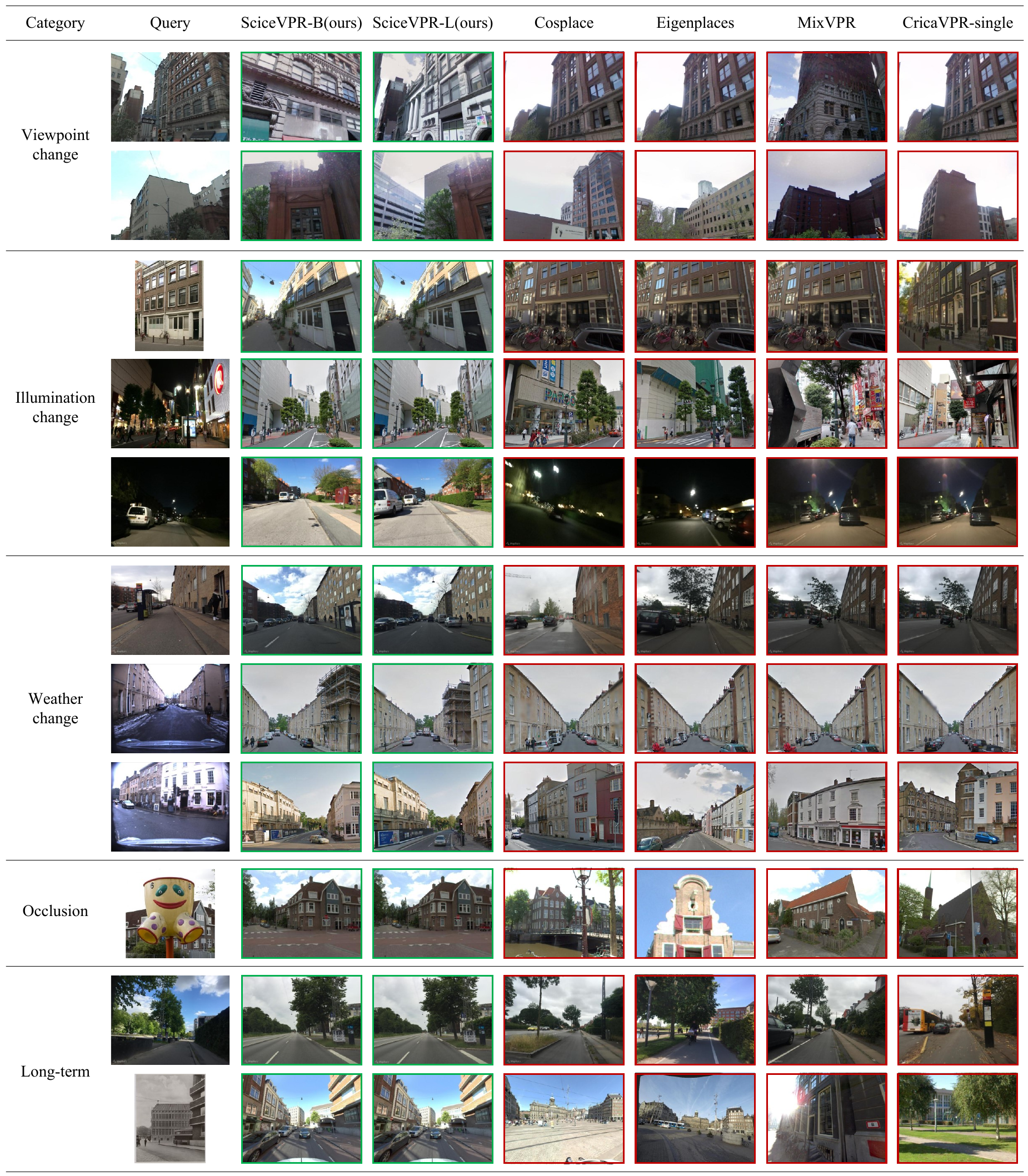}
\caption{Qualitative results of our SciceVPR models and the SOTA one-stage models in various challenging cases are shown. The SciceVPR models correctly recognize the true place for all the listed distinguishing queries, while the other one-stage models fail.}
\label{fig5}
\end{figure*}

Furthermore, we compare our one-stage model SciceVPR-L, having a backbone similar to that of SelaVPR, with the SOTA two-stage models. As shown in Table \ref{tab:table2}, our SciceVPR-L outperforms all the two-stage models on the benchmark datasets except SelaVPR on the MSLS-val/challenge sets. The two-stage models re-rank the top-100 candidates after the global retrieval stage, whereas SciceVPR-L retrieves images solely based on global descriptors. Nevertheless, SciceVPR-L still yields competitive results and is on par with SelaVPR on the challenging MSLS-val and MSLS-challenge test sets. On the Pitts30k and Tokyo24/7 datasets, SciceVPR-L outperforms SelaVPR by 0.6\% and 3.1\% in Recall@1, demonstrating its advantages over existing SOTA models.

To further evaluate the generalization ability of SciceVPR across multiple domains, we compare SciceVPR with the SOTA one-stage models on AmsterTime \cite{amstertime} and SVOX \cite{svox} datasets. As shown in Table \ref{tab:table3}, SciceVPR is more robust than one-stage models when facing extreme weather variations and image modality changes (RGB database compared with gray query). To be specific, for Recall@1, SciceVPR-B scores 9.4\%, 11.5\% and 6.6\% higher than CricaVPR-single on AmsterTime, SVOX-Night and SVOX-Sun datasets, respectively. As well, SciceVPR-L performs better than one-stage models by a large margin. Overall, SciceVPR is effective in dealing with VPR tasks.

We also qualitatively present some scenarios difficult for VPR models to retrieve the correct results. The challenging examples include severe viewpoint changes, illumination changes between day and night, weather changes over the year, occlusions of buildings and the long-term structural variations at the same location. The top-1 retrieval results in Fig. \ref{fig5} illustrate that our SciceVPR models are robust enough to handle these tough queries and correctly identify their locations, while other models get confused by similar images that are far away from the queries.

\begin{figure*}[!t]
\centering
\includegraphics[width=6.5in]{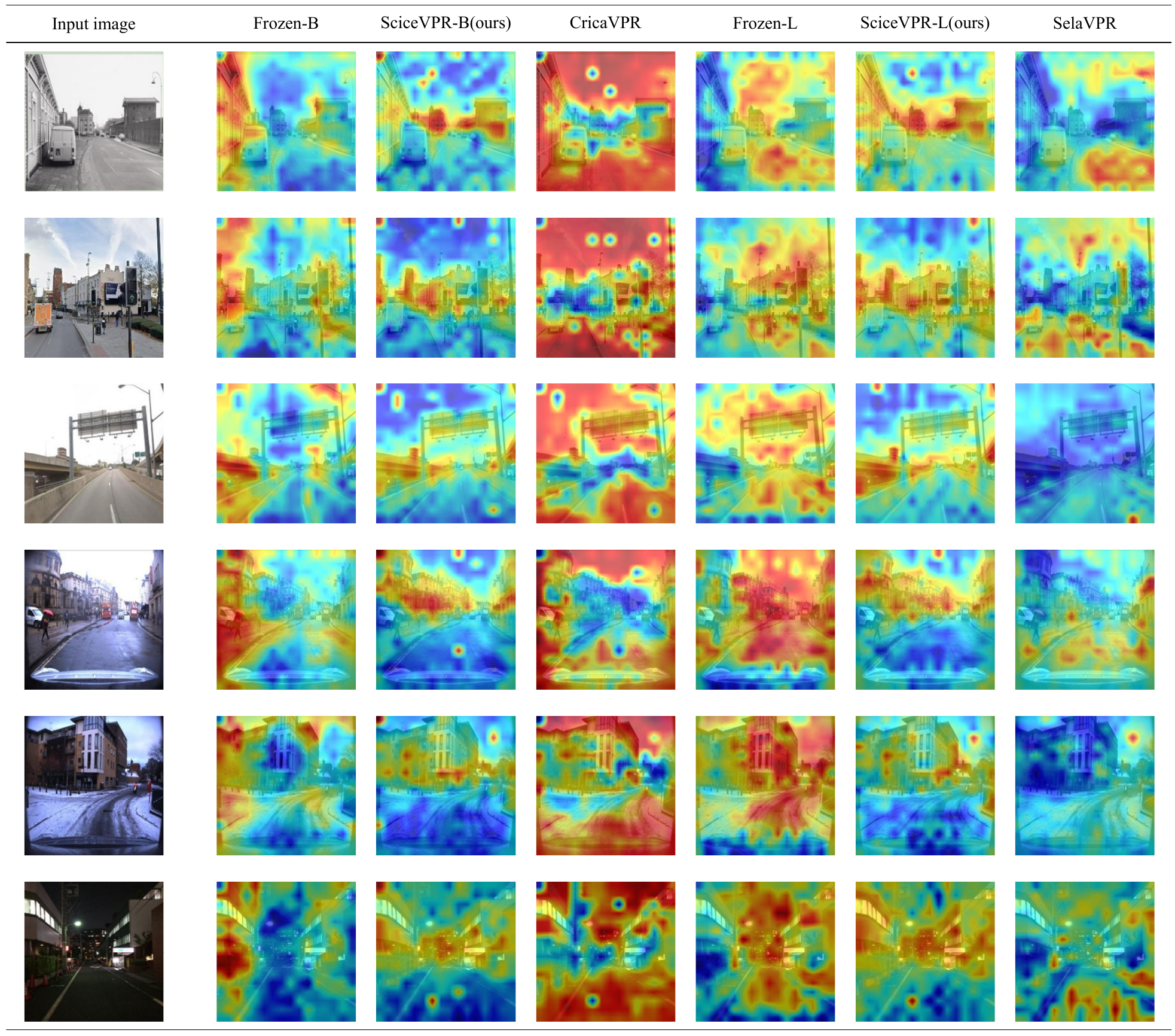}
\caption{The feature (attention) map visualizations. The feature maps of Frozen-B/L come from the concatenated features of the last 4 layers of the pre-trained DINOv2-B/L backbones. Similarly, we visualize the feature maps of the final output of the adapted DINOv2-B/L in CricaVPR/SelaVPR. Additionally, the feature maps of SciceVPR-B/L are from the conv(1,1) output. The heatmaps are generated by computing the mean across the channel dimension of the aforementioned feature maps, which are resized to $224 \times 224$. The heatmaps suggest that our models focus more on invariant features.}
\label{fig6}
\end{figure*}

\vspace{0.5em}
\noindent \textbf{Attention maps visualization.} To better understand the superior results of our SciceVPR models, we visualize the feature maps of the concatenated features of the last 4 layers of the pre-trained DINOv2-B/L, the final output of the adapted DINOv2-B/L in CricaVPR/SelaVPR and the output of conv(1,1) in our SciceVPR-B/L in Fig. \ref{fig6}. The heatmaps reveal that conv(1,1) learns to determine the task-invariant channel information, which focuses more on buildings regardless of variations caused by weather, day/night or image modality. In contrast, the concatenated output of features from the frozen pre-trained DINOv2-B/L models does not show such properties. On the other hand, the adapted DINOv2-B/L in CricaVPR/SelaVPR seems to focus on the sky and road, which may lead to incorrect retrieval results.

\subsection{Ablation Studies}
In the following ablation studies, we first validate the effectiveness of the proposed multi-layer feature fusion module and cross-image knowledge distillation, where the ablated models no longer use PCA for dimensionality reduction. Then, we conduct experiments to determine the impact of feature dimensions on the results. All the SciceVPR variants share the same multi-layer feature fusion module structure as the distilled Super-CricaVPR with only the parameters of conv(1,1) being passed to SciceVPR and then frozen unless specified otherwise.

\begin{table}[]
\caption{Ablation study on the multi-layer feature fusion module with the best results in \textbf{bold}. \label{tab:table4}}
\centering
\footnotesize
\setlength{\tabcolsep}{3pt}
\begin{spacing}{1.25}
\begin{tabular}{c|cc|cc|cc}
\Xhline{1pt}
\multirow{2}{*}{Method} & \multicolumn{2}{c|}{Pitts30k-test} & \multicolumn{2}{c|}{Tokyo24/7} & \multicolumn{2}{c}{MSLS-challenge} \\ \cline{2-7} 
                        & R@1              & R@5             & R@1            & R@5           & R@1              & R@5             \\ \hline
CricaVPR-single         & 91.6             & 95.7            & 89.5           & 94.0          & 66.9             & 79.3            \\
Super-CricaVPR-B-single & \textbf{92.2}    & \textbf{96.5}   & \textbf{93.3}  & \textbf{96.8} & \textbf{67.9}    & \textbf{82.8}   \\ \Xhline{1pt}
\end{tabular}
\end{spacing}
\end{table}

\vspace{0.5em}
\noindent \textbf{Ablation study on the multi-layer feature fusion module.} We compare standard Super-CricaVPR-B-single with CricaVPR-single to demonstrate the advantages of our multi-layer feature fusion module. The only difference between the two is in the feature extraction stage. Table \ref{tab:table4} shows that the Super-CricaVPR-B-single model appears to have better performance than that of the CricaVPR-single model on multiple datasets, confirming that using frozen multi-layer DINOv2 features with learnable channel-wise and spatial fusion layers can produce more suitable local features for the aggregation module on VPR tasks than making adaptations to DINOv2. 

\begin{figure}[!t]
\centering
\includegraphics[width=3.4in]{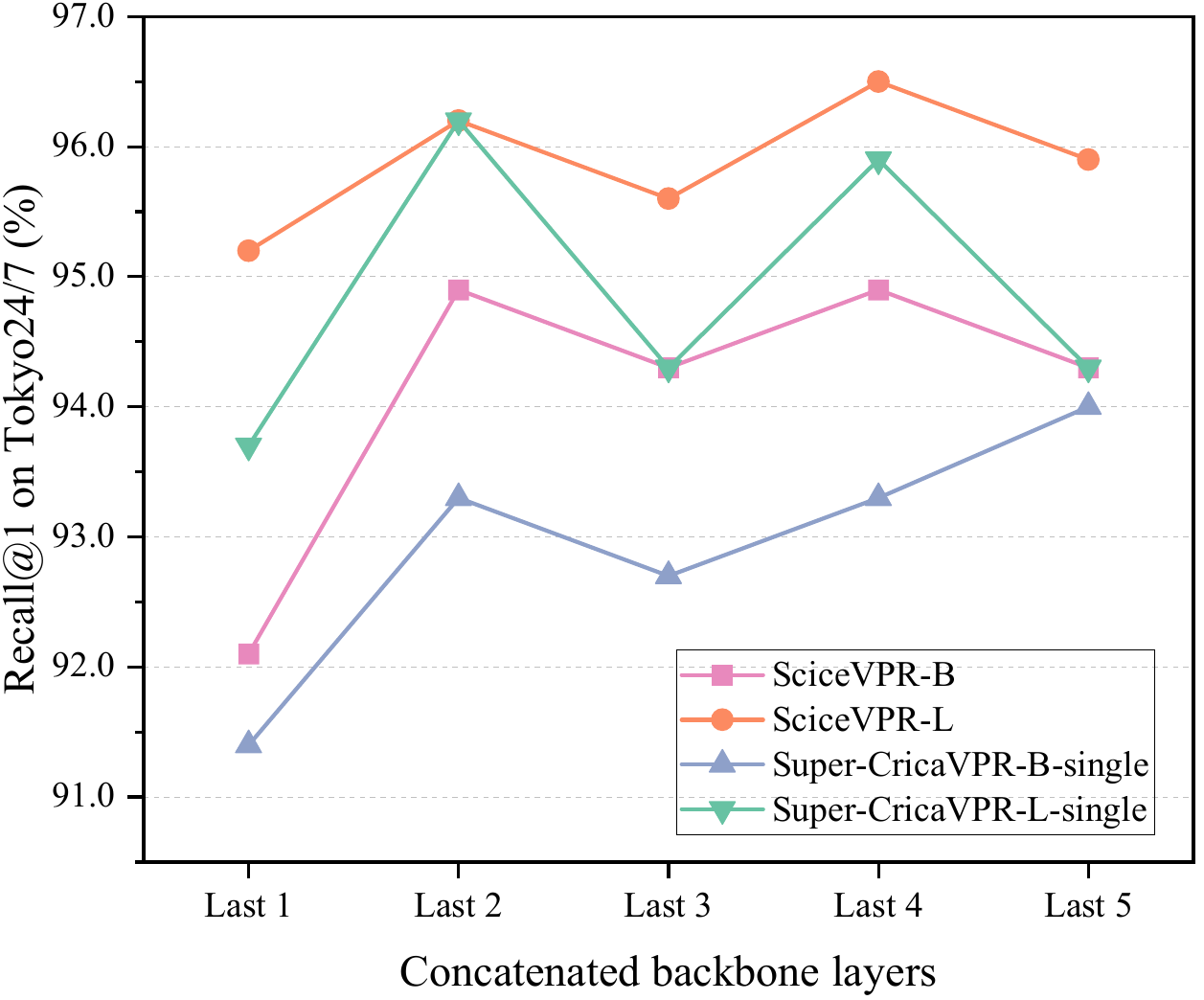}
\caption{Test results of different models with different concatenated backbone layers on Tokyo24/7.}
\label{fig7}
\end{figure}

\begin{figure}[!t]
\centering
\subfloat[]{\includegraphics[width=1.7in]{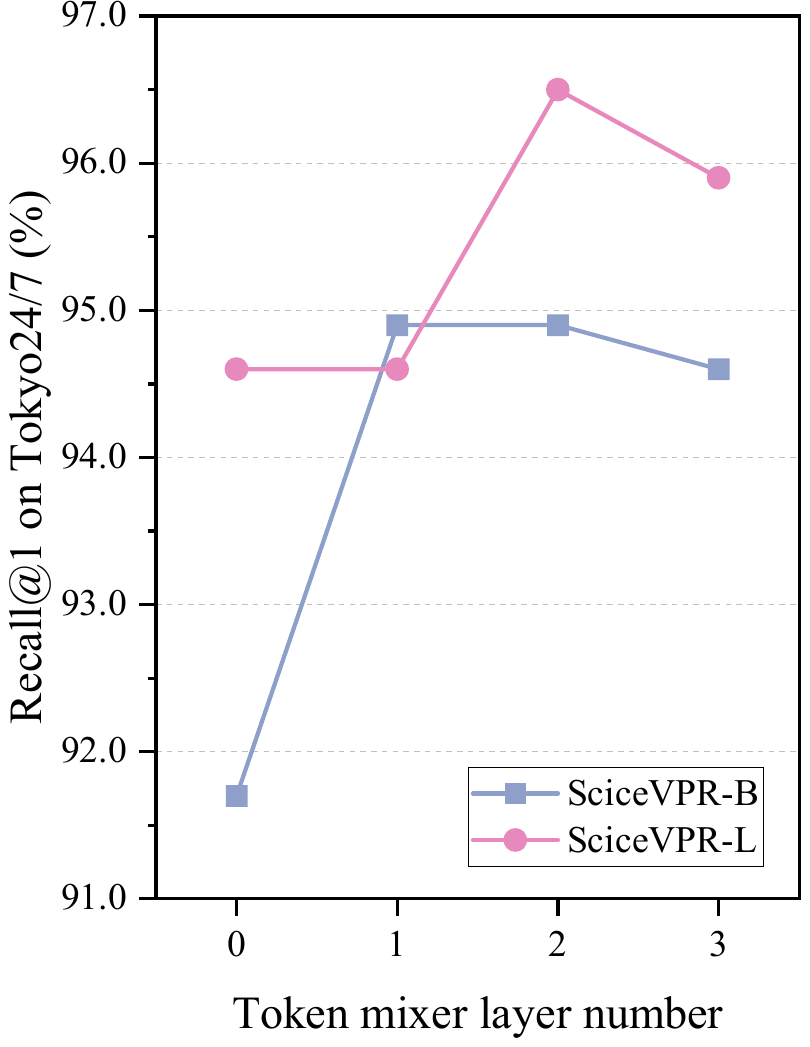}%
\label{fig_first_case_1}}
\hfil
\subfloat[]{\includegraphics[width=1.7in]{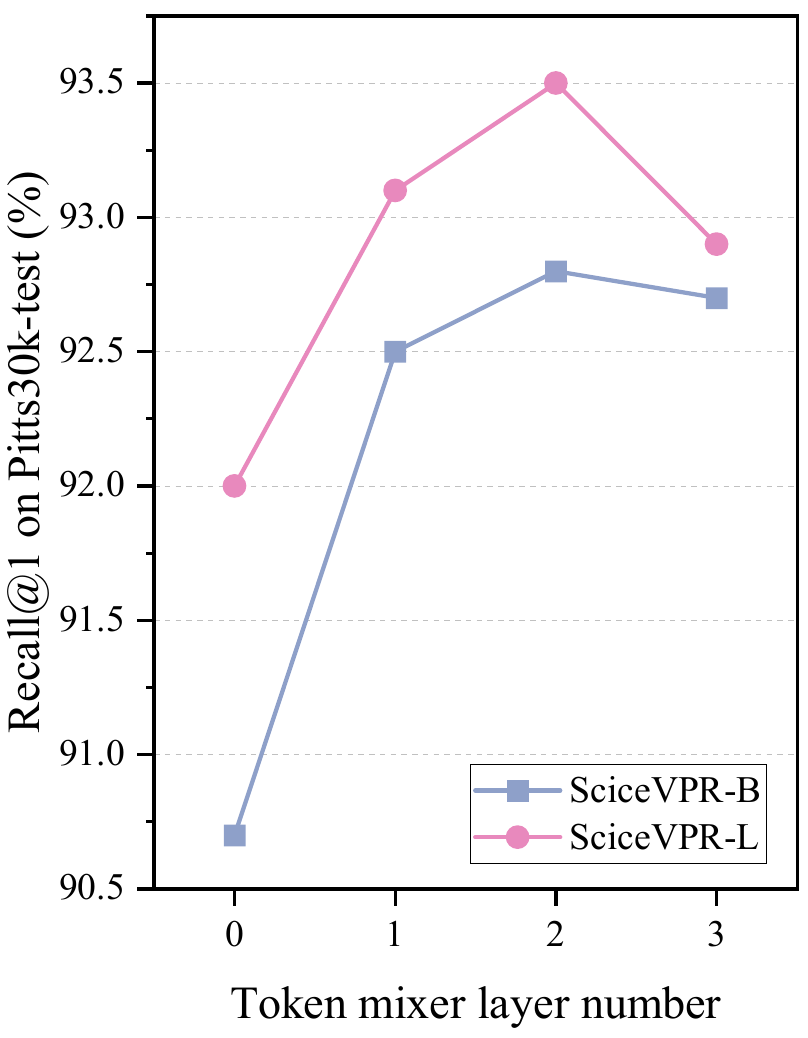}%
\label{fig_second_case_1}}
\caption{Test results of SciceVPR-B and SciceVPR-L models with different number of token mixer layers on (a) Tokyo24/7 and on (b) Pitts30k-test.}
\label{fig8}
\end{figure}

To understand better each component of our multi-layer feature fusion module, we investigate the appropriate number of concatenated backbone layers and token mixer layers, as displayed in Fig. \ref{fig7} and Fig. \ref{fig8}, respectively. Firstly, we fix the token mixer layers (2 by default) and adjust the concatenated layers among the trained Super-CricaVPR models together with the distilled SciceVPR models. Results in Fig. \ref{fig7} make it clear that for SciceVPR models, it is better to utilize the last 4 concatenated backbone layers, regardless of whether the corresponding Super-CricaVPR models reach the peak indicated by the broken lines. Fig. \ref{fig7} also shows that SciceVPR always has better performance than that of Super-CricaVPR models with single input, demonstrating that our proposed self-enhanced encoder can successfully learn valuable cross-image information. Since the best architecture for Super-CricaVPR does not determine the best architecture for SciceVPR, we visualize only the results for SciceVPR in the following discussions. We fix the concatenated layers (4 by default) and change the number of token mixer layers on SciceVPR models. We can infer from Fig. \ref{fig8} that 2 is the best number of the token mixer layers, as it achieves better results than the others, when considering both Recall@1 on Tokyo24/7 and Pitts30k-test.

\begin{figure}[!t]
\centering
\subfloat[]{\includegraphics[width=1.7in]{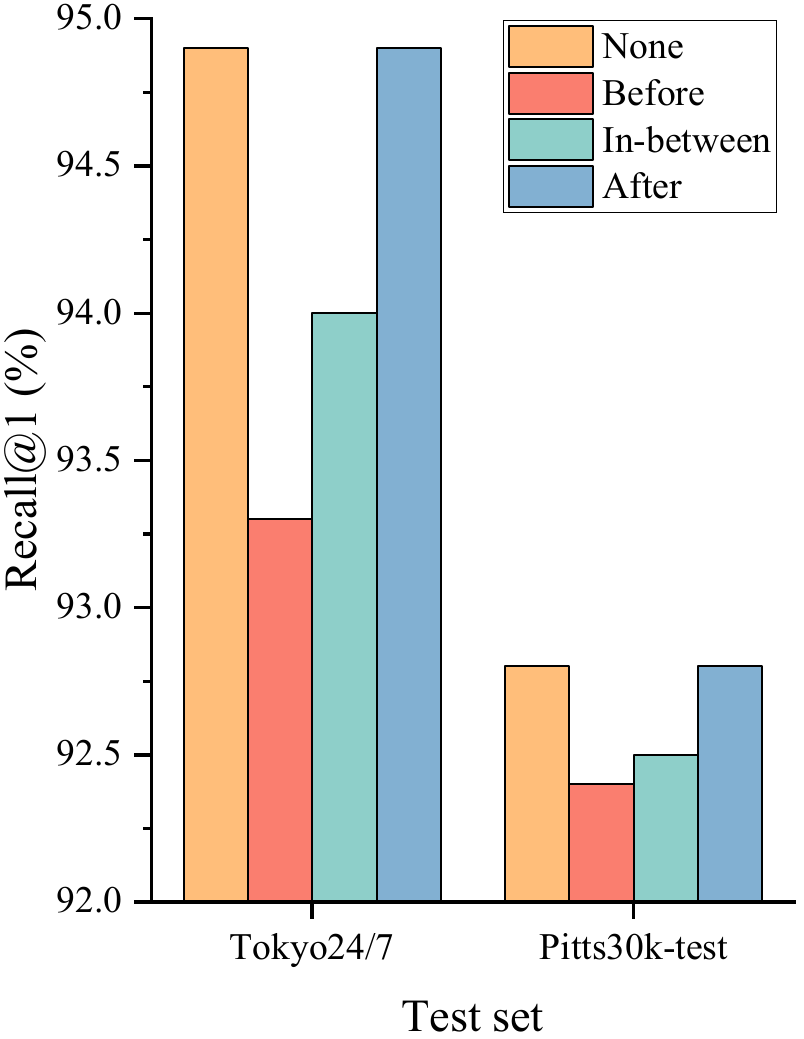}%
\label{fig_first_case_2}}
\hfil
\subfloat[]{\includegraphics[width=1.7in]{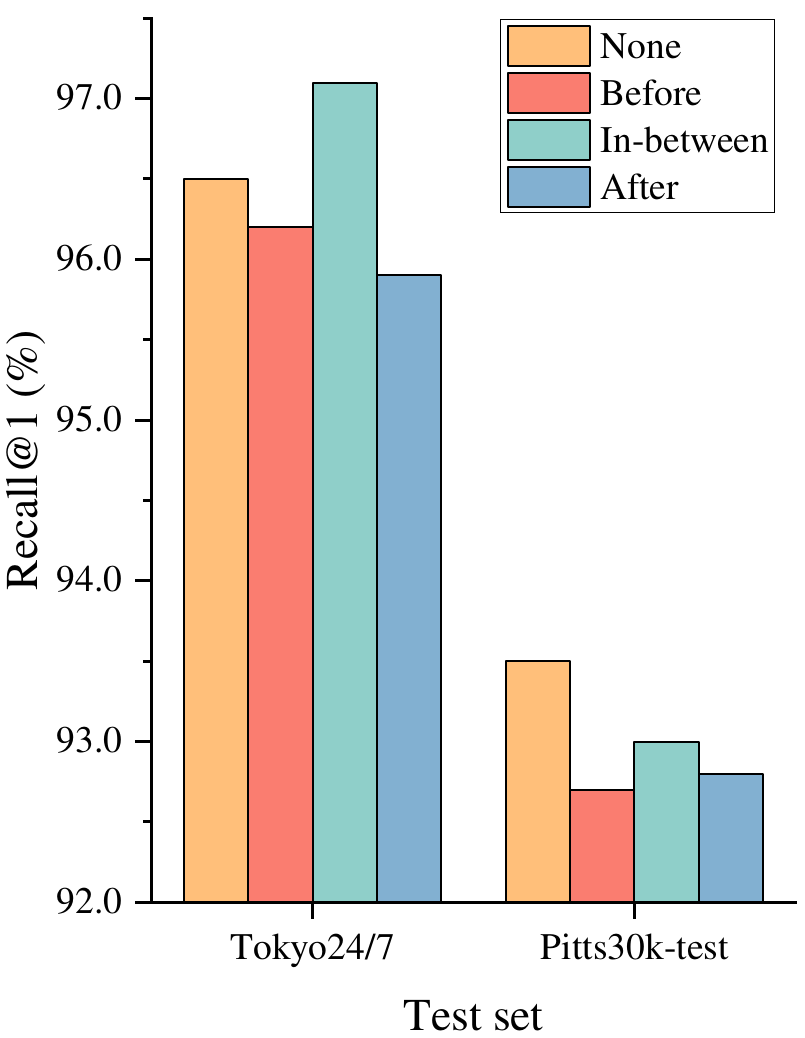}%
\label{fig_second_case_2}}
\caption{Test results of (a) SciceVPR-B and (b) SciceVPR-L models with an additional channel mixer layer before, in-between or after the token mixer layers, compared with the models without the channel mixer layer.}
\label{fig9}
\end{figure}

Interestingly, we find that incorporating a channel mixer layer into our multi-layer feature fusion module yields no improvement (see Fig. \ref{fig9}). In MLP-Mixer \cite{mlp-mixer}, a custom layer consists of one token mixer block and one channel mixer block, whereas we place a channel mixer layer before, in-between, and after our token mixer layers. The results in Fig. \ref{fig9} demonstrate that wherever there is an extra channel mixer layer, there is no improvement in performance. For example, SciceVPR-B with a channel mixer layer after the token mixer layers has the same Recall@1 with the standard SciceVPR-B. SciceVPR-L with a channel mixer in-between the token mixers scores 0.6\% higher in Recall@1 than that of the standard SciceVPR-L on Tokyo24/7, while it is 0.5\% lower in Recall@1 than that of the standard SciceVPR-L on Pitts30k-test. On the other hand, the task-relevant channel information from the concatenated DINOv2 layers has been fully explored using the $1 \times 1$ convolution and the multi-layer features from DINOv2 provide sufficient information to be explored. What remains to be investigated is the task-relevant features for spatial information, which are realized using the token mixer layers. 

\vspace{0.5em}
\noindent \textbf{Ablation study on the cross-image knowledge distillation.} The cross-image information implicitly encoded by our self-enhanced encoder is another important property that can improve the performance of SciceVPR, which is acquired through knowledge distillation from Super-CricaVPR. To evaluate the effectiveness of knowledge distillation, we compare SciceVPR models trained with or without the distillation loss. In particular, we transfer the parameters of conv(1,1) from Super-CricaVPR to SciceVPR, which are critical for selecting task-relevant channel information, and freeze them during the training of SciceVPR. Thus we experiment with and without passing the weights of conv(1,1). Once passed, the weights are frozen. During training, we find that SciceVPR models with distillation, i.e., using the passed weights of conv(1,1), converge faster. Hence, we train the SciceVPR-B models for 1 epoch and the SciceVPR-L models for 2 epochs, consistent with the standard SciceVPR implementation. SciceVPR models without distillation, i.e., without passing the weights of conv(1,1), are trained for 10 epochs, following the Super-CricaVPR implementation. 

\begin{table}[!ht]
\caption{Ablation study on the knowledge distillation and conv(1,1) pass with the best results in \textbf{bold}. \label{tab:table5}}
\centering
\footnotesize
\setlength{\tabcolsep}{2.5pt}
\begin{spacing}{1.25}
\begin{tabular}{c|c|c|cl|cl|cl}
\Xhline{1pt}
\multirow{2}{*}{Method}     & \multirow{2}{*}{\begin{tabular}[c]{@{}c@{}}Knowledge\\ distillation\end{tabular}} & \multirow{2}{*}{\begin{tabular}[c]{@{}c@{}}conv(1,1)\\ pass\end{tabular}} & \multicolumn{2}{c|}{Tokyo24/7}           & \multicolumn{2}{c|}{Pitts30k-test}       & \multicolumn{2}{c}{MSLS-val}            \\ \cline{4-9} 
                            &                                                                              &                                                                           & \multicolumn{2}{c|}{R@1}                 & \multicolumn{2}{c|}{R@1}                 & \multicolumn{2}{c}{R@1}                 \\ \hline
\multirow{4}{*}{SciceVPR-B} & \Checkmark                                                                            & \Checkmark                                                                         & \multicolumn{2}{c|}{\textbf{94.9(+0.0)}} & \multicolumn{2}{c|}{\textbf{92.8(+0.0)}} & \multicolumn{2}{c}{\textbf{89.2(+0.0)}} \\
                            & \Checkmark                                                                            & \XSolidBrush                                                                         & \multicolumn{2}{c|}{94.6(-0.3)}          & \multicolumn{2}{c|}{92.6(-0.2)}          & \multicolumn{2}{c}{88.8(-0.4)}          \\
                            & \XSolidBrush                                                                            & \Checkmark                                                                         & \multicolumn{2}{c|}{94.0(-0.9)}          & \multicolumn{2}{c|}{92.2(-0.6)}          & \multicolumn{2}{c}{88.4(-0.8)}          \\
                            & \XSolidBrush                                                                            & \XSolidBrush                                                                         & \multicolumn{2}{c|}{94.6(-0.3)}          & \multicolumn{2}{c|}{92.2(-0.6)}          & \multicolumn{2}{c}{88.2(-1.0)}          \\ \hline
\multirow{4}{*}{SciceVPR-L} & \Checkmark                                                                            & \Checkmark                                                                         & \multicolumn{2}{c|}{96.5(+0.0)}          & \multicolumn{2}{c|}{\textbf{93.5(+0.0)}} & \multicolumn{2}{c}{\textbf{90.7(+0.0)}} \\
                            & \Checkmark                                                                            & \XSolidBrush                                                                         & \multicolumn{2}{c|}{97.1(+0.6)}          & \multicolumn{2}{c|}{\textbf{93.5(+0.0)}} & \multicolumn{2}{c}{\textbf{90.7(+0.0)}} \\
                            & \XSolidBrush                                                                            & \Checkmark                                                                         & \multicolumn{2}{c|}{\textbf{97.5(+1.0)}} & \multicolumn{2}{c|}{93.3(-0.2)}          & \multicolumn{2}{c}{89.5(-1.2)}          \\
                            & \XSolidBrush                                                                            & \XSolidBrush                                                                         & \multicolumn{2}{c|}{94.6(-1.9)}          & \multicolumn{2}{c|}{93.2(-0.3)}          & \multicolumn{2}{c}{89.7(-1.0)}          \\ \Xhline{1pt}
\end{tabular}
\end{spacing}
\end{table}

\begin{table}[!ht]
\caption{Ablation study on the number of layers in the self-enhanced encoder with the best results in \textbf{bold}. \label{tab:table6}}
\centering
\footnotesize
\setlength{\tabcolsep}{2.5pt}
\begin{spacing}{1.25}
\begin{tabular}{c|c|cc|cc|cc}
\Xhline{1pt}
\multirow{2}{*}{Method}     & \multirow{2}{*}{\begin{tabular}[c]{@{}c@{}}encoder layer\\ number\end{tabular}} & \multicolumn{2}{c|}{Tokyo24/7} & \multicolumn{2}{c|}{Pitts30k-test} & \multicolumn{2}{c}{MSLS-val}  \\ \cline{3-8} 
                            &                                                                                 & R@1            & R@5           & R@1              & R@5             & R@1           & R@5           \\ \hline
\multirow{2}{*}{SciceVPR-B} & 1                                                                               & \textbf{94.9}  & \textbf{97.8} & \textbf{92.8}    & \textbf{96.8}   & 89.2          & \textbf{95.0} \\
                            & 2                                                                               & \textbf{94.9}  & \textbf{97.8} & 92.6             & 96.7            & \textbf{89.3} & \textbf{95.0} \\ \hline
\multirow{2}{*}{SciceVPR-L} & 1                                                                               & \textbf{96.5}  & \textbf{98.1} & \textbf{93.5}    & \textbf{96.9}   & 90.7          & 95.9          \\
                            & 2                                                                               & 96.2           & \textbf{98.1} & 93.4             & \textbf{96.9}   & \textbf{90.9} & \textbf{96.1} \\ \Xhline{1pt}
\end{tabular}
\end{spacing}
\end{table}

\begin{table}[!ht]
\caption{Ablation study on the number of dimensions of descriptors with the best results in \textbf{bold}. \label{tab:table7}}
\centering
\footnotesize
\setlength{\tabcolsep}{2.5pt}
\begin{spacing}{1.25}
\begin{tabular}{c|c|cc|cc|cc}
\Xhline{1pt}
\multirow{2}{*}{Method}     & \multirow{2}{*}{Dim} & \multicolumn{2}{c|}{Tokyo24/7} & \multicolumn{2}{c|}{Pitts30k-test} & \multicolumn{2}{c}{MSLS-val}  \\ \cline{3-8} 
                            &                      & R@1            & R@5           & R@1              & R@5             & R@1           & R@5           \\ \hline
\multirow{5}{*}{SciceVPR-B} & 10752                & \textbf{94.9}  & \textbf{97.8} & 92.8             & 96.8            & 89.2          & \textbf{95.0} \\
                            & 4096                 & \textbf{94.9}  & \textbf{97.8} & \textbf{92.9}    & 96.9            & \textbf{89.3} & \textbf{95.0} \\
                            & 2048                 & \textbf{94.9}  & \textbf{97.8} & 92.7             & \textbf{97.0}   & 88.8          & 94.9          \\
                            & 1024                 & 94.3           & \textbf{97.8} & 92.5             & 96.7            & 88.0          & 94.9          \\
                            & 512                  & 92.1           & 96.8          & 92.1             & 96.5            & 88.1          & 94.3          \\ \hline
\multirow{5}{*}{SciceVPR-L} & 10752                & 96.5           & \textbf{98.1} & \textbf{93.5}    & \textbf{96.9}   & \textbf{90.7} & \textbf{95.9} \\
                            & 4096                 & \textbf{97.1}  & \textbf{98.1} & 93.4             & \textbf{96.9}   & \textbf{90.7} & \textbf{95.9} \\
                            & 2048                 & \textbf{97.1}  & \textbf{98.1} & 93.1             & \textbf{96.9}   & 90.4          & \textbf{95.9} \\
                            & 1024                 & \textbf{97.1}  & \textbf{98.1} & 93.2             & 96.8            & 89.5          & 95.8          \\
                            & 512                  & 96.2           & 97.8          & 92.7             & 96.6            & 87.6          & 94.6          \\ \Xhline{1pt}
\end{tabular}
\end{spacing}
\end{table}

As shown in Table \ref{tab:table5}, SciceVPR models without distillation always exhibit worse Recall@1 performance compared to SciceVPR models with distillation, especially on the Tokyo24/7 test set, where the SciceVPR-L model trained with distillation surpasses the non-distilled variant by 1.9\% in Recall@1. Simply passing the conv(1,1) weights without distillation cannot boost the models' performance on all test sets, verifying that knowledge distillation contributes to the major improvement in performance. However, for the SciceVPR-B model with distillation, freezing the conv(1,1) weights from Super-CricaVPR-B helps to improve the model's performance. Hence, we choose to pass the weights in the implementation. We also experiment on the number of layers in the self-enhanced encoder with the standard training implementation and find that one layer is sufficient, as shown in Table \ref{tab:table6}.

\vspace{0.5em}
\noindent \textbf{Ablation study on the number of dimensions of the global descriptor.} Our models produce 10752-dimensional (10752-dim) global features, which may include some redundant or even noisy information. To address this issue, we perform PCA for dimensionality reduction and conduct an ablation study on the impact of the number of features' dimensions. As depicted in Table \ref{tab:table7}, 4096 is the optimal number of descriptor dimensions for both the SciceVPR-B and SciceVPR-L models, showing a slight advantage over the original 10752-dimensionality. It is also observed that our 2048-dim descriptors are comparable to the 4096-dim descriptors in terms of Recall@5 results across multiple test sets. These descriptors can serve as an appropriate substitute for the 4096-dim descriptors when high Recall@1 results are not required. In resource constrained situations, our 512-dim SciceVPR-B descriptors still surpass 512-dim CosPlace \cite{cosplace} descriptors by a large margin, and 512-dim SciceVPR-L descriptors perform on par with 2048-dim EigenPlaces \cite{eigenplaces} descriptors. This shows that our models are also competitive with low-dimensional descriptors (e.g., 512-dim).


\section{Conclusion}
In this paper, we propose a stable cross-image correlation enhanced model for visual place recognition called SciceVPR, which integrates the use of foundation models, feature fusion exploration, and contextual invariant information discovery to obtain robust and discriminative global descriptors. Firstly, the multi-layer feature fusion module of SciceVPR has an advantage over other VPR models in providing task-relevant local features. In this module, multi-layer features from DINOv2, which contains abundant visual representations, are concatenated and adjusted to include more task-relevant information through explicit channel and space fusion layers. Moreover, we distill the unstable cross-image correlation using a self-enhanced encoder in SciceVPR to obtain valuable cross-image invariant features resistant to VPR challenges. Extensive experiments on several datasets with diverse domain shifts establish that SciceVPR models can provide robust and discriminative global features and achieve new SOTA results among one-stage models with single input. 
Future work will focus on leveraging the foundation model and the cross-image correlation in cross-view geo-localization tasks, which are challenging due to drastic viewpoint changes between queries (e.g., satellite images) and database images (e.g., streetview images).

\bibliographystyle{IEEEtran}
\bibliography{reference}

@INPROCEEDINGS{netvlad,
  author={Arandjelovic, Relja and Gronat, Petr and Torii, Akihiko and Pajdla, Tomas and Sivic, Josef},
  booktitle={2016 IEEE Conf. Comput. Vis. Pattern Recognit. (CVPR)}, 
  title={{NetVLAD: CNN} Architecture for Weakly Supervised Place Recognition}, 
  year={2016},
  volume={},
  number={},
  pages={5297-5307},
  doi={10.1109/CVPR.2016.572}}

@ARTICLE{anyloc,
  author={Keetha, Nikhil and Mishra, Avneesh and Karhade, Jay and Jatavallabhula, Krishna Murthy and Scherer, Sebastian and Krishna, Madhava and Garg, Sourav},
  journal={IEEE Robot. Automat. Lett.}, 
  title={{AnyLoc}: Towards Universal Visual Place Recognition}, 
  year={2024},
  volume={9},
  number={2},
  pages={1286-1293},
  doi={10.1109/LRA.2023.3343602}}

@ARTICLE{probabilistic,
  author={Xu, Ming and Snderhauf, Niko and Milford, Michael},
  journal={IEEE Robot. Automat. Lett.}, 
  title={Probabilistic Visual Place Recognition for Hierarchical Localization}, 
  year={2021},
  volume={6},
  number={2},
  pages={311-318},
  doi={10.1109/LRA.2020.3040134}}

@ARTICLE{orb,
  author={Mur-Artal, Raúl and Tardós, Juan D.},
  journal={IEEE Trans. Robot.}, 
  title={{ORB-SLAM2}: An Open-Source SLAM System for Monocular, Stereo, and RGB-D Cameras}, 
  year={2017},
  volume={33},
  number={5},
  pages={1255-1262},
  doi={10.1109/TRO.2017.2705103}}

@INPROCEEDINGS{scalable,
  author={Doan, Dzung and Latif, Yasir and Chin, Tat-Jun and Liu, Yu and Do, Thanh-Toan and Reid, Ian},
  booktitle={2019 IEEE/CVF Int. Conf. Comput. Vis. (ICCV)}, 
  title={Scalable Place Recognition Under Appearance Change for Autonomous Driving}, 
  year={2019},
  volume={},
  number={},
  pages={9318-9327},
  doi={10.1109/ICCV.2019.00941}}

@ARTICLE{training,
  author={Nie, Jiwei and Feng, Joe-Mei and Xue, Dingyu and Pan, Feng and Liu, Wei and Hu, Jun and Cheng, Shuai},
  journal={IEEE Trans. Intell. Transp. Syst.}, 
  title={A Training-Free, Lightweight Global Image Descriptor for Long-Term Visual Place Recognition Toward Autonomous Vehicles}, 
  year={2024},
  volume={25},
  number={2},
  pages={1291-1302},
  doi={10.1109/TITS.2023.3320489}}

@InProceedings{openibl,
author="Ge, Yixiao
and Wang, Haibo
and Zhu, Feng
and Zhao, Rui
and Li, Hongsheng",
title="Self-supervising Fine-Grained Region Similarities for Large-Scale Image Localization",
booktitle="Comput. Vis. -- ECCV 2020",
year="2020",
pages="369--386",
isbn="978-3-030-58548-8"
}

@InProceedings{rerank1,
author="Cao, Bingyi
and Araujo, Andr{\'e}
and Sim, Jack",
title="Unifying Deep Local and Global Features for Image Search",
booktitle="Comput. Vis. -- ECCV 2020",
year="2020",
pages="726--743"
}

@INPROCEEDINGS{rerank2,
  author={Berton, Gabriele and Masone, Carlo and Paolicelli, Valerio and Caputo, Barbara},
  booktitle={2021 IEEE/CVF Int. Conf. Comput. Vis. (ICCV)}, 
  title={Viewpoint Invariant Dense Matching for Visual Geolocalization}, 
  year={2021},
  volume={},
  number={},
  pages={12149-12158},
  doi={10.1109/ICCV48922.2021.01195}}

@ARTICLE{gem,
  author={Radenović, Filip and Tolias, Giorgos and Chum, Ondřej},
  journal={IEEE Trans. Pattern Anal. Mach. Intell.}, 
  title={Fine-Tuning {CNN} Image Retrieval with No Human Annotation}, 
  year={2019},
  volume={41},
  number={7},
  pages={1655-1668},
  doi={10.1109/TPAMI.2018.2846566}}

@ARTICLE{pooling2,
  author={Nie, Jiwei and Xue, Dingyu and Pan, Feng and Ning, Zuotao and Liu, Wei and Hu, Jun and Cheng, Shuai},
  journal={IEEE Robot. Automat. Lett.}, 
  title={Efficient Saliency Encoding for Visual Place Recognition: Introducing the Lightweight Pooling-Centric Saliency-Aware {VPR} Method}, 
  year={2024},
  volume={9},
  number={7},
  pages={6035-6042},
  doi={10.1109/LRA.2024.3399591}}

@INPROCEEDINGS{loss1,
  author={Liu, Liu and Li, Hongdong and Dai, Yuchao},
  booktitle={2019 IEEE/CVF Int. Conf. Comput. Vis. (ICCV)}, 
  title={Stochastic Attraction-Repulsion Embedding for Large Scale Image Localization}, 
  year={2019},
  volume={},
  number={},
  pages={2570-2579},
  doi={10.1109/ICCV.2019.00266}}

@INPROCEEDINGS{loss2,
  author={Leyva-Vallina, María and Strisciuglio, Nicola and Petkov, Nicolai},
  booktitle={2023 IEEE/CVF Conf. Comput. Vis. Pattern Recognit. (CVPR)}, 
  title={Data-Efficient Large Scale Place Recognition with Graded Similarity Supervision}, 
  year={2023},
  volume={},
  number={},
  pages={23487-23496},
  doi={10.1109/CVPR52729.2023.02249}}

@INPROCEEDINGS{cosplace,
  author={Berton, Gabriele and Masone, Carlo and Caputo, Barbara},
  booktitle={2022 IEEE/CVF Conf. Comput. Vis. Pattern Recognit. (CVPR)}, 
  title={Rethinking Visual Geo-localization for Large-Scale Applications}, 
  year={2022},
  volume={},
  number={},
  pages={4868-4878},
  doi={10.1109/CVPR52688.2022.00483}}

@INPROCEEDINGS{eigenplaces,
  author={Berton, Gabriele and Trivigno, Gabriele and Caputo, Barbara and Masone, Carlo},
  booktitle={2023 IEEE/CVF Int. Conf. Comput. Vis. (ICCV)}, 
  title={{EigenPlaces}: Training Viewpoint Robust Models for Visual Place Recognition}, 
  year={2023},
  volume={},
  number={},
  pages={11046-11056},
  doi={10.1109/ICCV51070.2023.01017}}

@INPROCEEDINGS{mixvpr,
  author={Ali-Bey, Amar and Chaib-Draa, Brahim and Giguére, Philippe},
  booktitle={2023 IEEE/CVF Winter Conf. Appl. Comput. Vis. (WACV)}, 
  title={{MixVPR}: Feature Mixing for Visual Place Recognition}, 
  year={2023},
  volume={},
  number={},
  pages={2997-3006},
  doi={10.1109/WACV56688.2023.00301}}

@InProceedings{cricavpr,
    author    = {Lu, Feng and Lan, Xiangyuan and Zhang, Lijun and Jiang, Dongmei and Wang, Yaowei and Yuan, Chun},
    title     = {{CricaVPR}: Cross-image Correlation-aware Representation Learning for Visual Place Recognition},
    booktitle = {Proc. IEEE/CVF Conf. Comput. Vis. Pattern Recognit. (CVPR)},
    month     = {June},
    year      = {2024},
    pages     = {16772-16782}
}

@misc{dino-mix,
  title={{DINO-Mix}: Enhancing Visual Place Recognition with Foundational Vision Model and Feature Mixing}, 
  author={Gaoshuang Huang and Yang Zhou and Xiaofei Hu and Chenglong Zhang and Luying Zhao and Wenjian Gan and Mingbo Hou},
  year={2023},
  eprint={2311.00230},
  archivePrefix={arXiv},
  primaryClass={cs.CV},
  note={\emph{arXiv:2311.00230}},
}

@INPROCEEDINGS{transvpr,
  author={Wang, Ruotong and Shen, Yanqing and Zuo, Weiliang and Zhou, Sanping and Zheng, Nanning},
  booktitle={2022 IEEE/CVF Conf. Comput. Vis. Pattern Recognit. (CVPR)}, 
  title={{TransVPR}: Transformer-Based Place Recognition with Multi-Level Attention Aggregation}, 
  year={2022},
  volume={},
  number={},
  pages={13638-13647},
  doi={10.1109/CVPR52688.2022.01328}}

@INPROCEEDINGS{structvpr,
  author={Shen, Yanqing and Zhou, Sanping and Fu, Jingwen and Wang, Ruotong and Chen, Shitao and Zheng, Nanning},
  booktitle={2023 IEEE/CVF Conf. Comput. Vis. Pattern Recognit. (CVPR)}, 
  title={{StructVPR}: Distill Structural Knowledge with Weighting Samples for Visual Place Recognition}, 
  year={2023},
  volume={},
  number={},
  pages={11217-11226},
  doi={10.1109/CVPR52729.2023.01079}}

@INPROCEEDINGS{r2former,
  author={Zhu, Sijie and Yang, Linjie and Chen, Chen and Shah, Mubarak and Shen, Xiaohui and Wang, Heng},
  booktitle={2023 IEEE/CVF Conf. Comput. Vis. Pattern Recognit. (CVPR)}, 
  title={{$R^{2}$ Former}: Unified Retrieval and Reranking Transformer for Place Recognition}, 
  year={2023},
  volume={},
  number={},
  pages={19370-19380},
  doi={10.1109/CVPR52729.2023.01856}}

@inproceedings{selavpr,
  title={Towards Seamless Adaptation of Pre-trained Models for Visual Place Recognition},
  author={Lu, Feng and Zhang, Lijun and Lan, Xiangyuan and Dong, Shuting and Wang, Yaowei and Yuan, Chun},
  booktitle={12th Int. Conf. Learn. Represent.},
  year={2024}
}

@INPROCEEDINGS{transvlad,
  author={Xu, Yifan and Shamsolmoali, Pourya and Granger, Eric and Nicodeme, Claire and Gardes, Laurent and Yang, Jie},
  booktitle={2023 IEEE/CVF Winter Conf. Appl. Comput. Vis. (WACV)}, 
  title={{TransVLAD}: Multi-Scale Attention-Based Global Descriptors for Visual Geo-Localization}, 
  year={2023},
  volume={},
  number={},
  pages={2839-2848},
  doi={10.1109/WACV56688.2023.00286}}

@ARTICLE{hybrid,
  author={Wang, Yuwei and Qiu, Yuanying and Cheng, Peitao and Zhang, Junyu},
  journal={IEEE Trans. Circuits Syst. Video Technol.}, 
  title={Hybrid {CNN}-Transformer Features for Visual Place Recognition}, 
  year={2023},
  volume={33},
  number={3},
  pages={1109-1122},
  doi={10.1109/TCSVT.2022.3212434}}

@inproceedings{vgg16,
  title={Very deep convolutional networks for large-scale image recognition},
  author={Simonyan, K and Zisserman, A},
  booktitle={3rd Int. Conf. Learn. Represent.},
  year={2015}
}

@INPROCEEDINGS{resnet,
  author={He, Kaiming and Zhang, Xiangyu and Ren, Shaoqing and Sun, Jian},
  booktitle={2016 IEEE Conf. Comput. Vis. Pattern Recognit. (CVPR)}, 
  title={Deep Residual Learning for Image Recognition}, 
  year={2016},
  volume={},
  number={},
  pages={770-778},
  doi={10.1109/CVPR.2016.90}}

@misc{dinov2,
  title={{DINOv2}: Learning Robust Visual Features without Supervision}, 
  author={Maxime Oquab and Timothée Darcet and Théo Moutakanni and Huy Vo and Marc Szafraniec and Vasil Khalidov and Pierre Fernandez and Daniel Haziza and Francisco Massa and Alaaeldin El-Nouby and Mahmoud Assran and Nicolas Ballas and Wojciech Galuba and Russell Howes and Po-Yao Huang and Shang-Wen Li and Ishan Misra and Michael Rabbat and Vasu Sharma and Gabriel Synnaeve and Hu Xu and Hervé Jegou and Julien Mairal and Patrick Labatut and Armand Joulin and Piotr Bojanowski},
  year={2024},
  eprint={2304.07193},
  archivePrefix={arXiv},
  primaryClass={cs.CV},
  note={\emph{arXiv:2304.07193}},
}

@article{gsv-cities,
title = {{GSV-Cities}: Toward appropriate supervised visual place recognition},
journal = {Neurocomputing},
volume = {513},
pages = {194-203},
year = {2022},
issn = {0925-2312},
doi = {https://doi.org/10.1016/j.neucom.2022.09.127},
author = {Amar Ali-bey and Brahim Chaib-draa and Philippe Giguère}
}

@INPROCEEDINGS{msls,
  author={Warburg, Frederik and Hauberg, Søren and López-Antequera, Manuel and Gargallo, Pau and Kuang, Yubin and Civera, Javier},
  booktitle={2020 IEEE/CVF Conf. Comput. Vis. Pattern Recognit. (CVPR)}, 
  title={{Mapillary Street-Level Sequences}: A Dataset for Lifelong Place Recognition}, 
  year={2020},
  volume={},
  number={},
  pages={2623-2632},
  doi={10.1109/CVPR42600.2020.00270}}

@INPROCEEDINGS{bow,
  author={Philbin, James and Chum, Ondrej and Isard, Michael and Sivic, Josef and Zisserman, Andrew},
  booktitle={2007 IEEE Conf. Comput. Vis. Pattern Recognit. Recognition}, 
  title={Object retrieval with large vocabularies and fast spatial matching}, 
  year={2007},
  volume={},
  number={},
  pages={1-8},
  doi={10.1109/CVPR.2007.383172}}

@INPROCEEDINGS{vlad,
  author={Arandjelovic, Relja and Zisserman, Andrew},
  booktitle={2013 IEEE Conf. Comput. Vis. Pattern Recognit. Recognition}, 
  title={All About VLAD}, 
  year={2013},
  volume={},
  number={},
  pages={1578-1585},
  doi={10.1109/CVPR.2013.207}}

@article{sift,
author = {Lowe, David G.},
title = {Distinctive Image Features from Scale-Invariant Keypoints},
year = {2004},
volume = {60},
number = {2},
issn = {0920-5691},
doi = {10.1023/B:VISI.0000029664.99615.94},
journal = {Int. J. Comput. Vis.},
pages = {91-110},
numpages = {20}
}

@InProceedings{surf,
author="Bay, Herbert
and Tuytelaars, Tinne
and Van Gool, Luc",
title="{SURF}: Speeded Up Robust Features",
booktitle="Comput. Vis. -- ECCV 2006",
year="2006",
pages="404--417",
isbn="978-3-540-33833-8"
}

@INPROCEEDINGS{retrieval_vit,
  author={Gkelios, Socratis and Boutalis, Yiannis and Chatzichristofis, Savvas A.},
  booktitle={2021 17th Int. Conf. Distrib. Comput. Sensor Syst. (DCOSS)}, 
  title={Investigating the Vision Transformer Model for Image Retrieval Tasks}, 
  year={2021},
  volume={},
  number={},
  pages={367-373},
  doi={10.1109/DCOSS52077.2021.00065}}

@inproceedings{
vit,
title={An Image is Worth 16x16 Words: Transformers for Image Recognition at Scale},
author={Alexey Dosovitskiy and Lucas Beyer and Alexander Kolesnikov and Dirk Weissenborn and Xiaohua Zhai and Thomas Unterthiner and Mostafa Dehghani and Matthias Minderer and Georg Heigold and Sylvain Gelly and Jakob Uszkoreit and Neil Houlsby},
booktitle={9th Int. Conf. Learn. Represent.},
year={2021}
}

@misc{foundation_survey,
  title={On the opportunities and risks of foundation models}, 
  author={Rishi Bommasani and Drew A. Hudson and Ehsan Adeli and Russ Altman and Simran Arora and Sydney von Arx and Michael S. Bernstein and Jeannette Bohg and Antoine Bosselut and Emma Brunskill and Erik Brynjolfsson and others},
  year={2021},
  eprint={2108.07258},
  archivePrefix={arXiv},
  primaryClass={cs.CV},
  note={\emph{arXiv:2108.07258}}
}

@misc{distillation_survey1,
      title={Teacher-Student Architecture for Knowledge Distillation: A Survey}, 
      author={Chengming Hu and Xuan Li and Dan Liu and Haolun Wu and Xi Chen and Ju Wang and Xue Liu},
      year={2023},
      eprint={2308.04268},
      archivePrefix={arXiv},
      primaryClass={cs.LG},
      note={\emph{arXiv:2308.04268}}
}

@INPROCEEDINGS{mae,
  author={He, Kaiming and Chen, Xinlei and Xie, Saining and Li, Yanghao and Dollár, Piotr and Girshick, Ross},
  booktitle={2022 IEEE/CVF Conf. Comput. Vis. Pattern Recognit. (CVPR)}, 
  title={Masked Autoencoders Are Scalable Vision Learners}, 
  year={2022},
  volume={},
  number={},
  pages={15979-15988},
  doi={10.1109/CVPR52688.2022.01553}}

@INPROCEEDINGS{dino,
  author={Caron, Mathilde and Touvron, Hugo and Misra, Ishan and Jegou, Hervé and Mairal, Julien and Bojanowski, Piotr and Joulin, Armand},
  booktitle={2021 IEEE/CVF Int. Conf. Comput. Vis. (ICCV)}, 
  title={Emerging Properties in Self-Supervised Vision Transformers}, 
  year={2021},
  volume={},
  number={},
  pages={9630-9640},
  doi={10.1109/ICCV48922.2021.00951}}

@InProceedings{clip,
  title = 	 {Learning Transferable Visual Models From Natural Language Supervision},
  author =       {Radford, Alec and Kim, Jong Wook and Hallacy, Chris and Ramesh, Aditya and Goh, Gabriel and Agarwal, Sandhini and Sastry, Girish and Askell, Amanda and Mishkin, Pamela and Clark, Jack and Krueger, Gretchen and Sutskever, Ilya},
  booktitle = 	 {Proc. 38th Int. Conf. Mach. Learn.},
  pages = 	 {8748--8763},
  year = 	 {2021},
  volume = 	 {139}
}

@INPROCEEDINGS{foundation_downstream1,
  author={Pu, Nan and Zhong, Zhun and Sebe, Nicu},
  booktitle={2023 IEEE/CVF Conf. Comput. Vis. Pattern Recognit. (CVPR)}, 
  title={Dynamic Conceptional Contrastive Learning for Generalized Category Discovery}, 
  year={2023},
  volume={},
  number={},
  pages={7579-7588},
  doi={10.1109/CVPR52729.2023.00732}}

@ARTICLE{foundation_downstream2,
  author={Li, Jiaxing and Wong, Wai Keung and Jiang, Lin and Fang, Xiaozhao and Xie, Shengli and Xu, Yong},
  journal={IEEE Trans. Circuits Syst. Video Technol.}, 
  title={{CKDH: CLIP}-Based Knowledge Distillation Hashing for Cross-Modal Retrieval}, 
  year={2024},
  volume={34},
  number={7},
  pages={6530-6541},
  doi={10.1109/TCSVT.2024.3350695}}

@misc{distill_obj11,
      title={Distilling the Knowledge in a Neural Network}, 
      author={Geoffrey Hinton and Oriol Vinyals and Jeff Dean},
      year={2015},
      eprint={1503.02531},
      archivePrefix={arXiv},
      primaryClass={stat.ML},
      note={\emph{arXiv:1503.02531}}
}

@ARTICLE{distill_obj12,
  author={Zhang, Kangkai and Zhang, Chunhui and Li, Shikun and Zeng, Dan and Ge, Shiming},
  journal={IEEE Trans. Circuits Syst. Video Technol.}, 
  title={Student Network Learning via Evolutionary Knowledge Distillation}, 
  year={2022},
  volume={32},
  number={4},
  pages={2251-2263},
  doi={10.1109/TCSVT.2021.3090902}}

@ARTICLE{distill_obj13,
  author={Yang, Shunzhi and Xu, Liuchi and Zhou, Mengchu and Yang, Xiong and Yang, Jinfeng and Huang, Zhenhua},
  journal={IEEE Trans. Circuits Syst. Video Technol.}, 
  title={Skill-Transferring Knowledge Distillation Method}, 
  year={2023},
  volume={33},
  number={11},
  pages={6487-6502},
  doi={10.1109/TCSVT.2023.3271124}}

@INPROCEEDINGS{distill_obj21,
  author={Xie, Qizhe and Luong, Minh-Thang and Hovy, Eduard and Le, Quoc V.},
  booktitle={2020 IEEE/CVF Conf. Comput. Vis. Pattern Recognit. (CVPR)}, 
  title={Self-Training With Noisy Student Improves ImageNet Classification}, 
  year={2020},
  volume={},
  number={},
  pages={10684-10695},
  doi={10.1109/CVPR42600.2020.01070}}

@INPROCEEDINGS{distill_obj22,
  author={Xue, Zihui and Ren, Sucheng and Gao, Zhengqi and Zhao, Hang},
  booktitle={2021 IEEE/CVF Int. Conf. Comput. Vis. (ICCV)}, 
  title={Multimodal Knowledge Expansion}, 
  year={2021},
  volume={},
  number={},
  pages={834-843},
  doi={10.1109/ICCV48922.2021.00089}}

@ARTICLE{distill_obj31,
  author={Ren, Chuan-Xian and Ge, Pengfei and Yang, Peiyi and Yan, Shuicheng},
  journal={IEEE Trans. Neural Netw. Learn. Syst.}, 
  title={Learning Target-Domain-Specific Classifier for Partial Domain Adaptation}, 
  year={2021},
  volume={32},
  number={5},
  pages={1989-2001},
  doi={10.1109/TNNLS.2020.2995648}}

@ARTICLE{distill_obj32,
  author={Mei, Zhen and Ye, Peng and Li, Baopu and Chen, Tao and Fan, Jiayuan and Ouyang, Wanli},
  journal={IEEE Trans. Circuits Syst. Video Technol.}, 
  title={{DeNKD}: Decoupled Non-Target Knowledge Distillation for Complementing Transformer-Based Unsupervised Domain Adaptation}, 
  year={2024},
  volume={34},
  number={5},
  pages={3220-3231},
  doi={10.1109/TCSVT.2023.3315872}}

@INPROCEEDINGS{distill_obj41,
  author={Kundu, Jogendra Nath and Lakkakula, Nishank and Radhakrishnan, Venkatesh Babu},
  booktitle={2019 IEEE/CVF Int. Conf. Comput. Vis. (ICCV)}, 
  title={{UM-Adapt}: Unsupervised Multi-Task Adaptation Using Adversarial Cross-Task Distillation}, 
  year={2019},
  volume={},
  number={},
  pages={1436-1445},
  doi={10.1109/ICCV.2019.00152}}

@INPROCEEDINGS{distill_obj42,
  author={Ghiasi, Golnaz and Zoph, Barret and Cubuk, Ekin D. and Le, Quoc V. and Lin, Tsung-Yi},
  booktitle={2021 IEEE/CVF Int. Conf. Comput. Vis. (ICCV)}, 
  title={Multi-Task Self-Training for Learning General Representations}, 
  year={2021},
  volume={},
  number={},
  pages={8836-8845},
  doi={10.1109/ICCV48922.2021.00873}}

@article{distillation_survey2,
author = {Gou, Jianping and Yu, Baosheng and Maybank, Stephen J. and Tao, Dacheng},
title = {Knowledge Distillation: A Survey},
year = {2021},
volume = {129},
number = {6},
issn = {0920-5691},
doi = {10.1007/s11263-021-01453-z},
journal = {Int. J. Comput. Vision},
pages = {1789-1819},
numpages = {31}
}

@inproceedings{distill_feature1,
  author       = {Adriana Romero and
                  Nicolas Ballas and
                  Samira Ebrahimi Kahou and
                  Antoine Chassang and
                  Carlo Gatta and
                  Yoshua Bengio},
  title        = {{FitNets}: Hints for Thin Deep Nets},
  booktitle    = {3rd Int. Conf. Learn. Represent.},
  year         = {2015}
}

@INPROCEEDINGS{distill_relation1,
  author={Park, Wonpyo and Kim, Dongju and Lu, Yan and Cho, Minsu},
  booktitle={2019 IEEE/CVF Conf. Comput. Vis. Pattern Recognit. (CVPR)}, 
  title={Relational Knowledge Distillation}, 
  year={2019},
  volume={},
  number={},
  pages={3962-3971},
  doi={10.1109/CVPR.2019.00409}}

@ARTICLE{distill_relation2,
  author={Zhang, Kangkai and Ge, Shiming and Shi, Ruixin and Zeng, Dan},
  journal={IEEE Trans. Circuits Syst. Video Technol.}, 
  title={Low-Resolution Object Recognition With Cross-Resolution Relational Contrastive Distillation}, 
  year={2024},
  volume={34},
  number={4},
  pages={2374-2384},
  doi={10.1109/TCSVT.2023.3310042}}

@inproceedings{mlp-mixer,
 author = {Tolstikhin, Ilya O and Houlsby, Neil and Kolesnikov, Alexander and Beyer, Lucas and Zhai, Xiaohua and Unterthiner, Thomas and Yung, Jessica and Steiner, Andreas and Keysers, Daniel and Uszkoreit, Jakob and Lucic, Mario and Dosovitskiy, Alexey},
 booktitle = {Adv. Neural Inf. Process. Syst.},
 pages = {24261--24272},
 title = {{MLP-Mixer}: An all-{MLP} Architecture for Vision},
 volume = {34},
 year = {2021}
}

@INPROCEEDINGS{ms_loss,
  author={Wang, Xun and Han, Xintong and Huang, Weilin and Dong, Dengke and Scott, Matthew R.},
  booktitle={2019 IEEE/CVF Conf. Comput. Vis. Pattern Recognit. (CVPR)}, 
  title={Multi-Similarity Loss With General Pair Weighting for Deep Metric Learning}, 
  year={2019},
  volume={},
  number={},
  pages={5017-5025},
  doi={10.1109/CVPR.2019.00516}}

@INPROCEEDINGS{pitts30k,
  author={Torii, Akihiko and Sivic, Josef and Pajdla, Tomá and Okutomi, Masatoshi},
  booktitle={2013 IEEE Conf. Comput. Vis. Pattern Recognit.}, 
  title={Visual Place Recognition with Repetitive Structures}, 
  year={2013},
  volume={},
  number={},
  pages={883-890},
  doi={10.1109/CVPR.2013.119}}

@INPROCEEDINGS{tokyo,
  author={Torii, Akihiko and Arandjelović, Relja and Sivic, Josef and Okutomi, Masatoshi and Pajdla, Tomas},
  booktitle={2015 IEEE Conf. Comput. Vis. Pattern Recognit. (CVPR)}, 
  title={24/7 place recognition by view synthesis}, 
  year={2015},
  volume={},
  number={},
  pages={1808-1817},
  doi={10.1109/CVPR.2015.7298790}}

@INPROCEEDINGS{amstertime,
  author={Yildiz, Burak and Khademi, Seyran and Siebes, Ronald Maria and Van Gemert, Jan},
  booktitle={2022 26th Int. Conf. on Pattern Recognit. (ICPR)}, 
  title={{AmsterTime}: A Visual Place Recognition Benchmark Dataset for Severe Domain Shift}, 
  year={2022},
  volume={},
  number={},
  pages={2749-2755},
  doi={10.1109/ICPR56361.2022.9956049}}

@INPROCEEDINGS{svox,
  author={Moreno Berton, Gabriele and Paolicelli, Valerio and Masone, Carlo and Caputo, Barbara},
  booktitle={2021 IEEE Winter Conf. Appl. Comput. Vis. (WACV)}, 
  title={Adaptive-Attentive Geolocalization from few queries: a hybrid approach}, 
  year={2021},
  volume={},
  number={},
  pages={2917-2926},
  doi={10.1109/WACV48630.2021.00296}}

@article{oxford,
author = {Maddern, Will and Pascoe, Geoffrey and Linegar, Chris and Newman, Paul},
title = {1 year, 1000 km: The Oxford RobotCar dataset},
year = {2017},
volume = {36},
number = {1},
issn = {0278-3649},
journal = {Int. J. Rob. Res.},
pages = {3-15},
numpages = {13}
}



\end{document}